\documentclass[lettersize,journal]{IEEEtran}
\usepackage{amsmath,amsfonts}
\usepackage{algorithmic}
\usepackage{algorithm}
\usepackage{array}
\usepackage[caption=false,font=normalsize,labelfont=sf,textfont=sf]{subfig}
\usepackage{textcomp}
\usepackage{stfloats}
\usepackage{url}
\usepackage{verbatim}
\usepackage{graphicx}
\usepackage{cite}
\usepackage{booktabs}
\usepackage{subcaption}
\usepackage{multirow}
\usepackage{threeparttable}
\usepackage{makecell}
\usepackage{xcolor}
\usepackage{fp}
\begin{document}

\title{DIO: Refining Mutual Information and Causal Chain to Enhance Machine Abstract Reasoning Ability}

\author{Beiming Yuan, Ruizhuo Song, Member, IEEE
\thanks{This work was supported by the National Natural Science Foundation of China under Grants 62273036. Corresponding author: Ruizhuo Song, ruizhuosong@ustb.edu.cn}
\thanks{Ruizhuo Song and Beiming Yuan are with the Beijing Engineering Research Center of Industrial Spectrum Imaging, School of Automation and Electrical Engineering, University of Science and Technology Beijing, Beijing 100083, China (Ruizhuo Song email: ruizhuosong@ustb.edu.cn and Beiming Yuan email: d202310354@xs.ustb.edu.cn). }

}

\markboth{Journal of \LaTeX\ Class Files,~Vol.~14, No.~8, August~2021}%
{Shell \MakeLowercase{\textit{et al.}}: A Sample Article Using IEEEtran.cls for IEEE Journals}


\maketitle

\begin{abstract}

Despite the outstanding performance of deep learning across various domains, their fundamental bottleneck in abstract reasoning remains unresolved. To tackle this challenge, the community has introduced Raven’s Progressive Matrices (RPM) as an authoritative benchmark for evaluating abstract reasoning, pattern recognition, and complex problem-solving capabilities. Centering on RPM, this paper aims to enhance the abstract reasoning ability of machine intelligence.
This paper proposes a unified ``causal-information'' framework that systematically addresses three pivotal questions: (i) how to embed a human-like reasoning chain into network architecture; (ii) how to tighten the variational lower bound on mutual information, which limits existing models; and (iii) how to reduce semantic misalignment between learned representations and human-like causal chains.
We deliver four incremental contributions:
\begin{enumerate}
  \item {DIO} -- a baseline that explicitly encodes the full causal chain ``image $\rightarrow$ attributes $\rightarrow$ patterns $\rightarrow$ consistency $\rightarrow$ answer'' and reveals that its learning objective, like that of previous end-to-end RPM solvers, is essentially a variational lower bound on the mutual information between contexts and solutions, thereby exposing how the tightness of this bound governs the performance of such models;
  \item {Brando} -- identifies that one key opportunity for tightening this bound is to supply  additional ``constructive incorrect options,” and designs a learnable mapper to provide such options, offering a novel strategy for tightening the variational lower bound on mutual information in reasoning models;
  \item {WORLD} -- models RPM image features with a Gaussian mixture, enabling unlimited, diverse, and target-sampled incorrect options that push DIO toward the theoretical bound and empower open-ended {generative} RPM solving;
  \item {DIEGO} -- rectifies semantic biases in the ``attributes $\rightarrow$ patterns'' link, aligning internal representations with human causal frameworks and boosting out-of-distribution (o.o.d.) generalization.
\end{enumerate}
Experiments show that our complete model achieves significant performance gains on RAVEN, I-RAVEN, and PGM benchmarks, exhibits stronger generalization on several o.o.d. sub-tasks, and is capable of generating structurally diverse RPM solutions.
This study not only achieves remarkable performance in abstract visual reasoning but also validates ``causal-chain-modeling plus mutual-information-tightening'' as a core design principle for AI.


\end{abstract}


\section{Introduction}

\IEEEPARstart{D}{eep} learning, a cross-disciplinary technology inspired by the brain's operational mechanisms, has demonstrated formidable potential across numerous fields and achieved outstanding accomplishments in generative tasks. Key technologies, including Generative Adversarial Networks (GANs)\cite{GAN}, Variational Autoencoders (VAEs)\cite{VAE} and the attention-based Transformer architecture\cite{Transformer}, enable models to generate high-quality images, texts and videos from random noise or given conditions, thereby expanding AI's scope in content creation.

In CV, deep CNNs and their variants extract and classify features efficiently, surpassing traditional methods in detection and segmentation \cite{ResNet, A survey of convolutional neural networks, A survey of visual transformers}. In NLP, deep language models learn grammar, semantics, and context, excelling in translation and generation \cite{Transformer, GPT-3, Attention in natural language processing}. In multimodal VQA, fused visual-linguistic representations enable joint reasoning to answer natural-language questions about images \cite{VQA}.
Large Language Models (LLMs), with massive parameters and data, now dominate deep learning by delivering unified, high-performance NLP solutions and driving AI toward generality. Yet recent studies \cite{ LLM2, LLM3, LLM4} consistently show that LLMs encounter clear bottlenecks in abstract reasoning tasks that demand concept understanding, logical reasoning, and strategy formation, which are core hallmarks of human intelligence and key yardsticks for model capability. These shortcomings reveal current limits in tackling complex cognition and underscore the critical need to boost LLMs' abstract reasoning capacity for stronger, more general artificial intelligence.

This paper focuses on enhancing deep learning's abstract reasoning ability through effective model architecture design, training strategy optimization, and data augmentation methods that aim to provide both theory and practice to break current bottlenecks and extend deep learning to a broader perspective and cognitive tasks.


\section{Problem Description}

In AI, abstract reasoning is regarded as a core hallmark of intelligence and a key prerequisite for achieving human-level or superhuman AI. The research community has proposed Raven’s Progressive Matrices (RPM)\cite{RPM} as an evaluation benchmark, thereby presenting the challenge of designing algorithms that simultaneously handle symbolic logic, causal modeling, and uncertainty.

\subsection{Raven's Progressive Matrices}

Raven’s Progressive Matrices (RPM) \cite{RPM} constitute an authoritative benchmark for assessing higher-order cognitive abilities in deep-learning systems, targeting core facets of intelligence such as abstract reasoning, pattern recognition and complex problem solving. Their standardized protocol correlates strongly with human intelligence metrics, offering a reliable gauge of model cognitive complexity. Today, the RAVEN \cite{RAVENdataset} and PGM \cite{PGMdataset} datasets stand as the two most representative RPM testbeds.

\begin{figure}[htp]\centering
	\includegraphics[trim=3.5cm 0cm 2cm 0cm, clip, width=8.5
 cm]{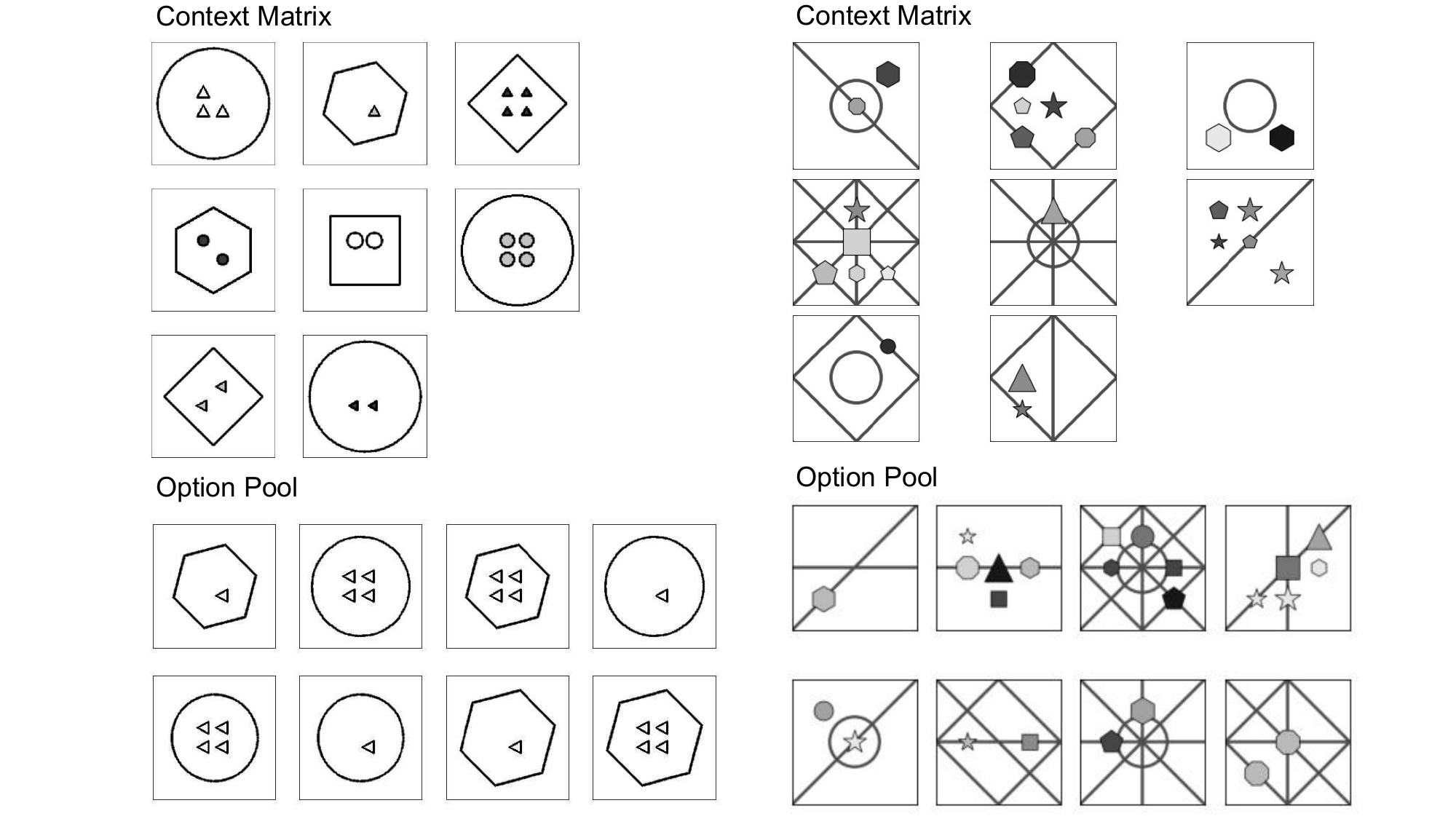}
	\caption{RAVEN and PGM Instance}
\label{RAVEN}
\end{figure}

The RAVEN dataset comprises structured RPM instances, with each test unit containing 16 images: 8 forming a ``context matrix" as the reasoning basis, and the remaining 8 constituting a candidate option pool. The task requires selecting appropriate images from the options to construct a complete 3$\times$3 matrix that adheres to progressive geometric transformation rules and abstract logical principles. Figure \ref{RAVEN} (left) illustrates a typical instance from the RAVEN dataset. Similarly, while the PGM dataset also employs 8 images for context and options, its underlying logical rules must satisfy dual constraints in both row and column directions of the matrix. Figure \ref{RAVEN} (right) presents a standard instance from the PGM dataset.

\subsection{Related work}

Image reasoning models such as CoPINet\cite{CoPINet} and DCNet\cite{DCNet} primarily focus on learning differences and rules. Meanwhile, models like NCD\cite{NCD}, SAVIR-T\cite{SAVIR-T}, as well as neuro-symbolic systems (including PrAE, NVSA, ALANS\cite{PrAE,ALANS,NVSA}), have achieved promising results in both enhancing model interpretability and improving reasoning accuracy. RS-CNN and RS-TRAN\cite{RS} demonstrate exceptional capabilities in addressing problems related to RPM.

\section{Methodology}

This paper first proposes that to enhance the ability of deep learning models to solve RPM, which are causal reasoning problems, when designing the model's structure and dividing its modules, one should fully refer to the causal chains within the problems. Based on this methodology, this paper designs a new model for solving RPM problems, named DIO. However, this paper finds that DIO is in satin without substance. Its learning approach is essentially the same as that of the previous RPM-solving model, whose purpose is to maximize the mutual information between the context matrix and the correct option in the RPM instance. This learning approach leads to a situation where, no matter how well its structure aligns with the causal chains in the problems, its learning and optimization will not progress towards the human predefined reasoning framework.
Therefore, this paper proposes three methods, namely Brando, WORLD, and DIEGO, to refine DIO's learning objectives. As a bonus, when DIO adopts the WORLD method, it demonstrates the capability to solve generative RPM problems.

\begin{figure}[ht]\centering
	\includegraphics[trim=0cm 0cm 0cm 0cm, clip, width=8.5cm]{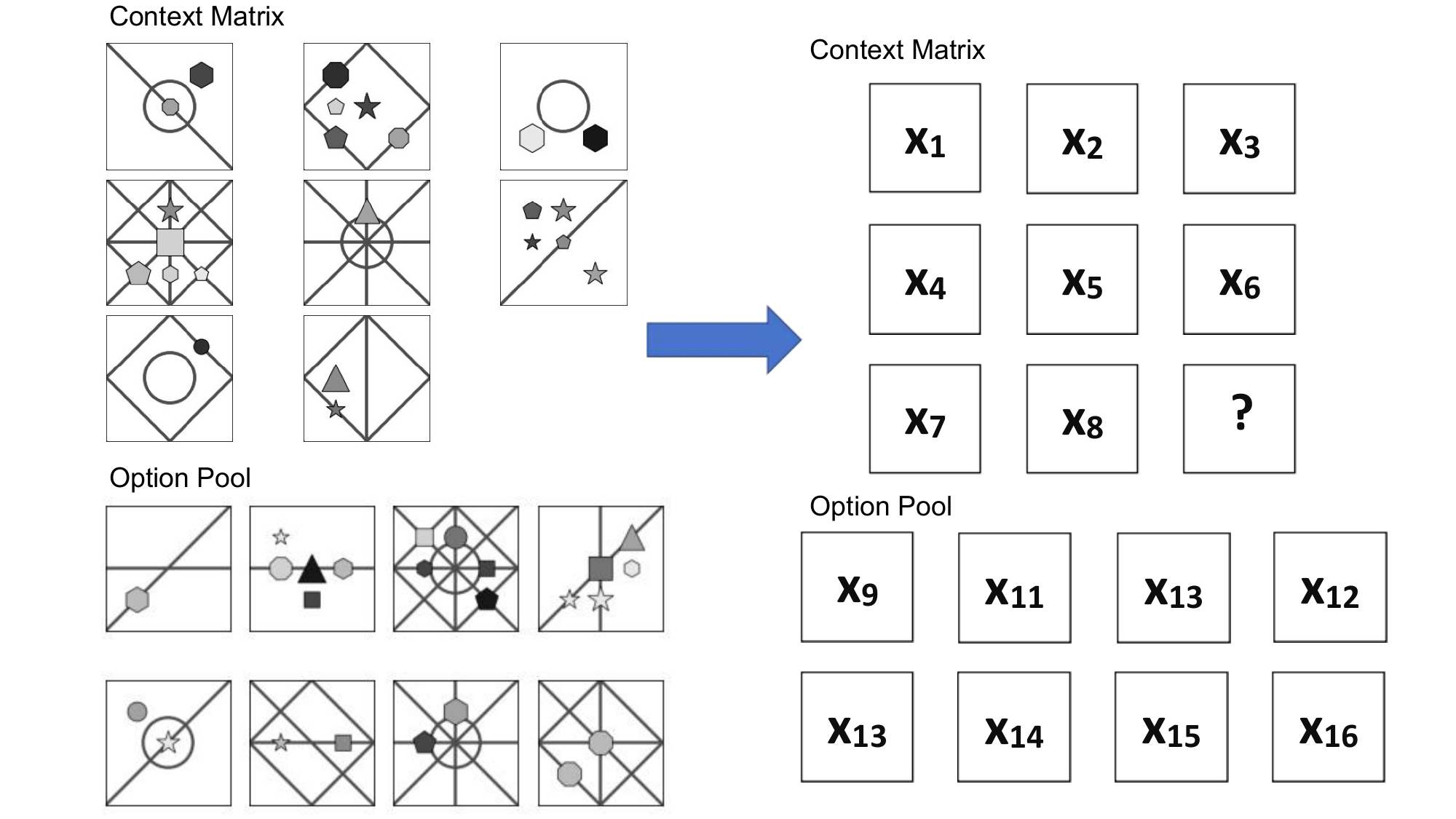}
	\caption{Annotations of Images Within an RPM Instance.}
\label{annotates}
\end{figure}

The RPM instances in this paper use the annotation approach illustrated in Figure~\ref{annotates}, which allows for a clearer explanation of our model and methods.
Specifically, we denote the 16 RPM images as $\{x_i | i \in [1, 16]\}$.

\section{DIO: A Baseline Model}

In this section, we design a high-performance RPM-solving model, named DIO (Deep-learning Intelligent-model Organized via Causal Chains). While providing a detailed exposition of the model's structure, we have also elucidated its underlying design principles.

\subsection{The Causal Chains in RPM Problems.}

The process of solving RPM problems proceeds as follows: the model must first identify the progressive patterns governing abstract image attributes across the context matrices $\{x_i | i \in [1,8]\}$, and then select the correct option $x_\alpha$, which comprehensively continues these patterns~\cite{RPMInductivebias}, from the option pool $\{x_i | i \in [9,16]\}$.
This implies that there exists a causal chain in RPM problems: RPM images $\rightarrow$ abstract attributes $\rightarrow$ progressive patterns of attributes $\rightarrow$ consistency of progressive patterns $\rightarrow$ correctness of options. 

This paper proposes that designing the structure of an RPM-solving model in accordance with this causal chain will yield high performance, which prompted us to design the DIO model. 
Accordingly, the DIO comprises four modules: the image feature extraction module, the progressive pattern induction module, the progressive pattern consistency evaluation module, and the option evaluation module.

\subsection{The Image Feature Extraction Module.} 

\begin{figure}[htp]\centering
	\includegraphics[trim=0cm 3cm 0cm 1.5cm, clip, width=6.5
 cm]{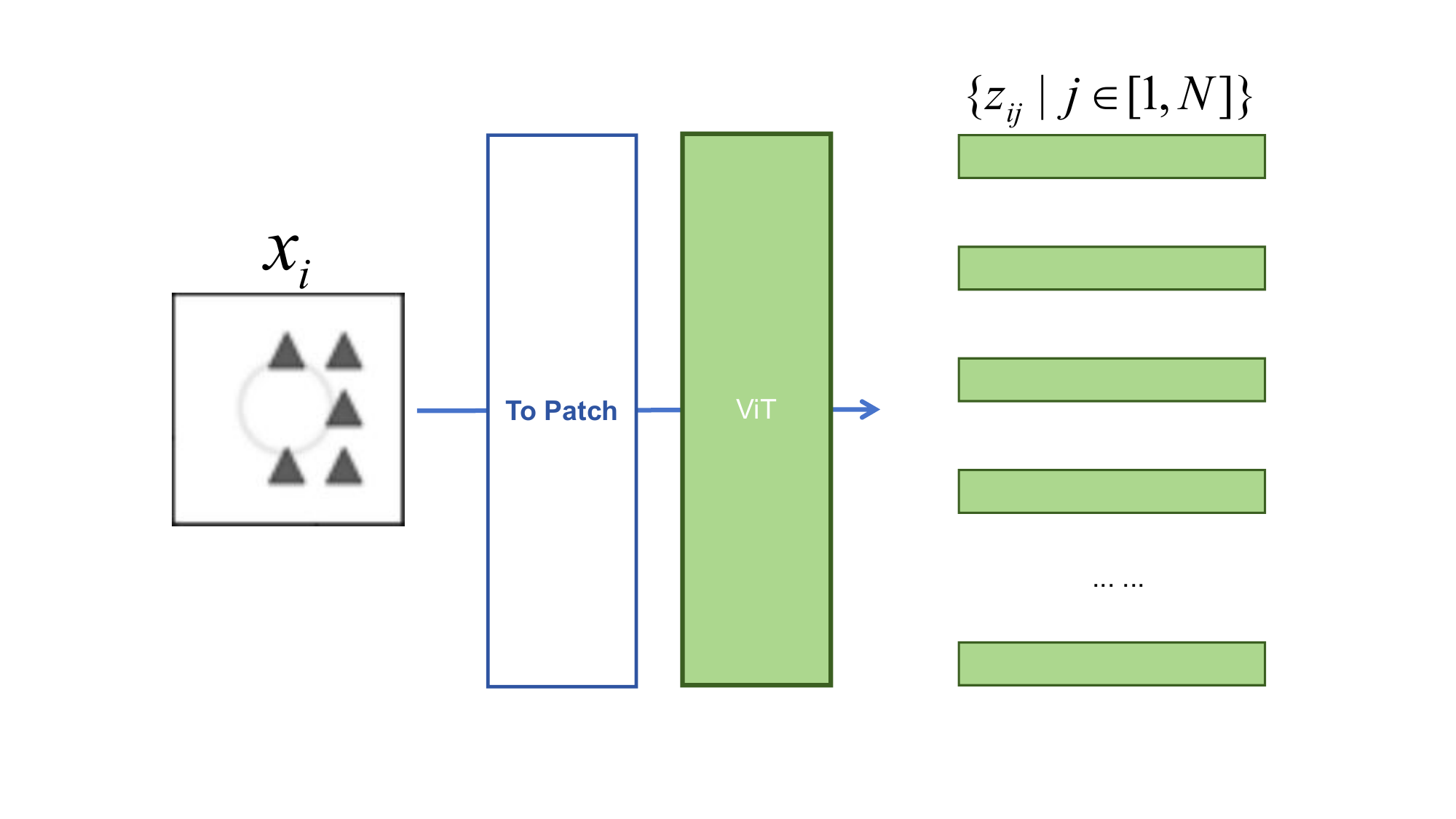}
	\caption{The Image Feature Extraction Module.}
\label{Feature Extraction Module}
\end{figure}

RPM problems place significant emphasis on evaluating the abstract reasoning capabilities of deep models. Consequently, when extracting features from RPM images, we opt for a Visual Transformer (ViT) \cite{ViT} that prioritizes image context. Figure \ref{Feature Extraction Module} illustrates the process by which ViT extracts the tokenized image features \( \{z_{ij}|j \in [1,N]\} \) from the input image \( x_i \), where \( j \) denotes the token index of the image features and $N$ denotes the total patch count.
This process can be formally expressed as:
\begin{equation}\label{eq1}
    \{z_{ij}|i\in[1,16], j \in [1,N]\} = \text{FEM}(x_i)|_{i=1}^{16}
\end{equation}


\subsection{The Progressive Pattern Induction Module}

\begin{figure}[htp]\centering
	\includegraphics[trim=0cm 0cm 0cm 0cm, clip, width=7.5
 cm]{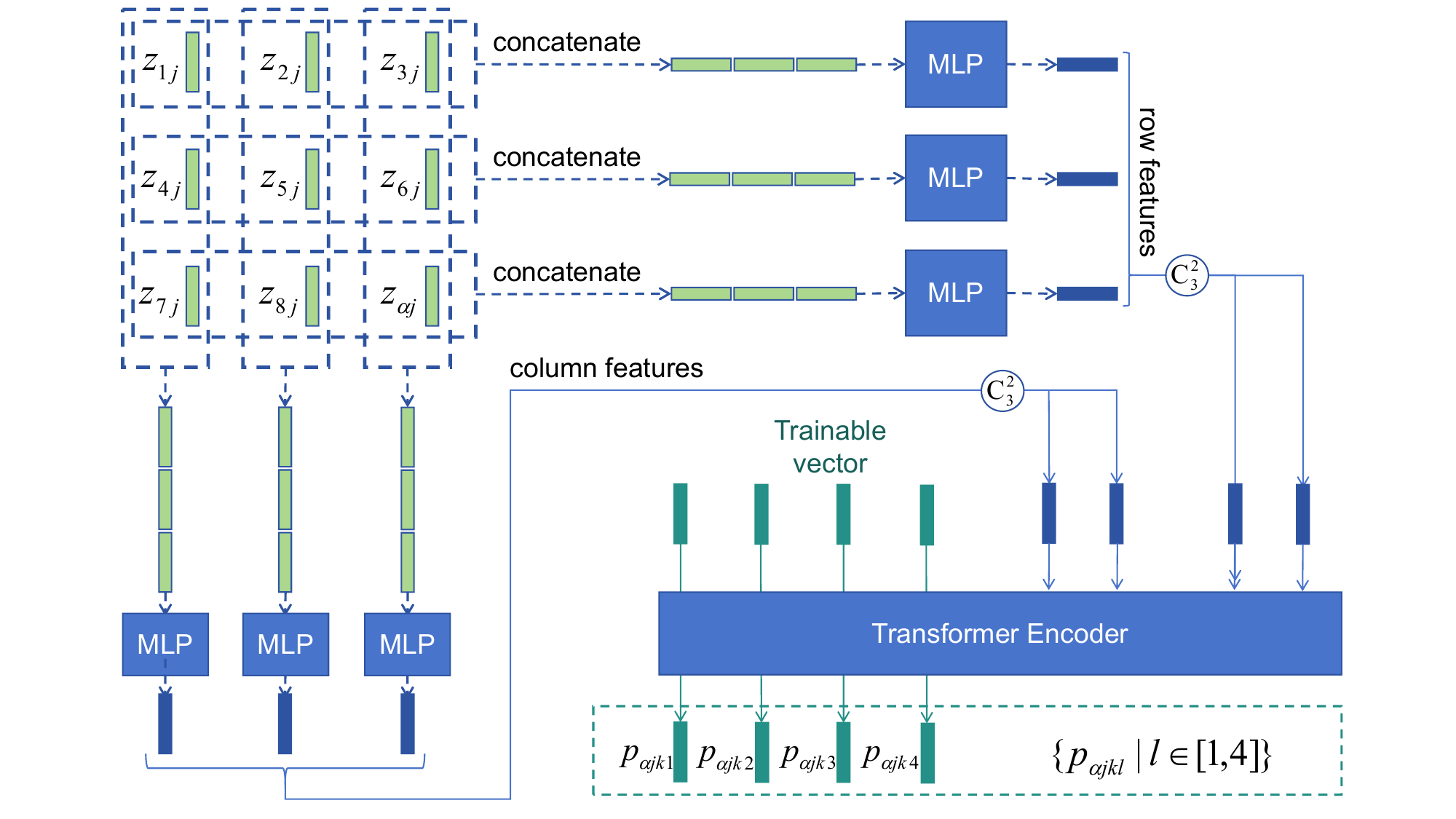}
	\caption{The Progressive Pattern Induction Module.}
\label{The Progressive Pattern Induction Module}
\end{figure}

The deliberate design of RPM problems necessitates that participants must simultaneously observe two rows and two columns to identify the progressive patterns within the matrix. This represents a critical segment in the causal chain of RPM problems. Consequently, the progressive pattern induction module of DIO adopts the structure illustrated in Figure \ref{The Progressive Pattern Induction Module}.
Figure \ref{The Progressive Pattern Induction Module} indicates that after appending the option features $\{z_{\alpha j} | j \in [1, N]\}$ into the context matrix $\{z_{ij} |i \in [1, 8], j \in [1, N]\}$, DIO summarizes the progressive patterns among the features that share the same token index $j$. First, DIO utilizes a Multi-Layer Perceptron (MLP) to extract row and column features from the matrix. Subsequently, it randomly selects two row features from the three available row features and two column features from the three column features, combining these four features for further processing.
After binding four trainable vectors of the same dimension to the selected four features, DIO employs a Transformer Encoder to process them into eight logical outputs. Among the output vector groups, the four vectors corresponding to the trainable vectors are denoted as $\{P_{\alpha jl} |l \in [1,4]\}$.

$\{P_{\alpha jl} | l \in [1,4]\}$ are regarded as the partial representations of progressive patterns within the matrix $\{z_{ij}, z_{\alpha j} |i \in [1, 8]\}$. If all $C_{3}^{2} \cdot C_{3}^{2} = 9$ possible row–column selections are considered (where $C_n^r$ denotes the number of combinations of $n$ items taken $r$ at a time), we obtain comprehensive representations $\{P_{\alpha jkl} | k \in [1,9], l \in [1,4]\}$ of these patterns.
This module can be formally expressed as:
\begin{align}\label{PPIM}
    &\{P_{\alpha jkl}|j\in [1,N],k\in[1,9],l\in[1,4]\}\nonumber\\ 
    &= \text{PPIM}(\{z_{ij},z_{\alpha j}| i\in[1,8]\})|_{j=1}^N
\end{align}

\subsection{The Progressive Pattern Consistency Evaluation Module}

As its name suggests, DIO's progressive pattern consistency evaluation module is responsible for using a Transformer Encoder to compute the consistency among progressive-pattern representations $P_{\alpha jkl}$ that share the same $j$-index and $l$-index.
Specifically, the Transformer Encoder performs a dimension-preserving mapping on $\{P_{\alpha jkl} | k \in [1,9]\}$ to obtain $\{P'_{\alpha jkl} | k \in [1,9]\}$. Repeating this for all $j\in[1,N]$ and $l\in[1,4]$ yields the complete representation set $\{P'_{\alpha jkl} |j\in[1,N], k \in [1,9], l\in[1,4]\}$. Finally, each vector in this full set is mapped to a scalar via an MLP, producing the consistency score $s_{\alpha jkl}$.
This entire module can be represented as:
\begin{align}
 \{s_{\alpha jkl} |j\in[1,N], k \in [1,9],l\in [1,4]\}\nonumber\\ 
= \text{PCEM}(\{P_{\alpha jkl} | k \in [1,9]\})|_{j=1,l=1}^{N,4}
\end{align}

\subsection{The Option Evaluation Module.}
DIO's option evaluation module computes the score \( S_\alpha \) that reflects the correctness of option \(x_ \alpha \) by averaging the consistency scores $\{s_{\alpha jkl} |j\in[1,N],k \in [1,9],l\in [1,4]\}$:
\begin{align}\label{eq6}
S_\alpha &= \frac{1}{N\cdot 9\cdot 4}\big(\sum_{j=1}^{N} \sum_{k=1}^{9} \sum_{l=1}^{4} s_{\alpha jkl}\big)
\end{align}
Subsequently, the softmax function maps the scores $\{S_i \mid i \in [8,16]\}$ derived by DIO to a probability distribution $\hat P(x \mid \{x_i\}_{i=1}^8,\theta)$ parameterized by DIO:
\begin{align}\label{eq7}
    \hat P(x = x_\alpha|\{x_i\}_{i=1}^8,\theta ) = \frac{e^{S_\alpha}}{\sum_{{\hat \alpha} = 9}^{16}e^{S_{{\hat \alpha}}}}
\end{align}
We introduce a loss function ${\ell}_\text{DIO}$ to refine this parameterized distribution $\hat P(x|\{x_i\}_{i=1}^8,\theta)$, thereby enabling end-to-end training of all parameters across DIO's modules:
\begin{align}\label{loss}
{\ell}_\text{DIO} = \mathbb{E}\left[\log (\frac{\sum_{\substack{{\hat \alpha} = 9\\{\hat \alpha}\neq \alpha}}^{16} \hat{P}( x_{\hat{\alpha}} | \{x_i\}_{i=1}^8, \theta)}{\hat{P}(x_{\alpha} | \{x_i\}_{i=1}^8, \theta)} + \delta)
\right]-\log \delta
\end{align}
where $\mathbb{E}[\cdot]$ denotes the expectation over the data distribution; $\delta$ is a small safeguard against numerical overflow, initialized at 1 and annealed to $1\times10^{-5}$ as training progresses.


\section{Deviation and Potential Correction Methods for Learning Objectives of End-to-End RPM-solving Model}

In this section, we provide a theoretical analysis anchored in DIO’s optimization objective to assess the validity of the intuitively motivated methodology of causal-chain-based architecture optimization.

\subsection{The Learning Objective of DIO}
By abstracting away the specific architecture of DIO (i.e., by collapsing Equations~(\ref{eq1})--(\ref{eq6})), we obtain a unified formulation that highlights its core inductive mechanism:
\begin{align} 
    S_\alpha&= \text{DIO}(\{x_i,x_\alpha|i\in[1,8]\}) \label{soloDIO}
\end{align}
Using Equation (\ref{eq7}) and \eqref{soloDIO}, we recast the parameterized distribution $\hat P(x_\alpha | \{x_i\}_{i=1}^8,\theta)$ into a form that explicitly embeds DIO's inductive mechanism :
\begin{align}\label{eq10}
    \hat P(x_\alpha|\{x_i\}_{i=1}^8,\theta)= \frac{e^{\text{DIO}(\{x_i,x_\alpha|i\in[1,8]\})}}{\sum_{{\hat \alpha} = 9}^{16}e^{\text{DIO}(\{x_i,x_{\hat \alpha}|i\in[1,8]\})}}
\end{align}
Accordingly, the loss function $\ell_\text{DIO}$ can be rewritten as:
\begin{align}\label{eq102}
    \ell_{\text{DIO}}= \mathbb{E}\left[\log\frac{\sum_{\substack{{\hat \alpha} = 9\\{\hat \alpha}\neq \alpha}}^{16}e^{\text{DIO}(\{x_i,x_{\hat \alpha}|i\in[1,8]\})}}{e^{\text{DIO}(\{x_i,x_\alpha|i\in[1,8]\})}}\right]
\end{align}
For the sake of discussion, we have temporarily removed the safeguard $\delta$.

Notably, \(I(x_\alpha; \{x_i\}_{i=1}^8)\), which denotes the mutual information between the context matrix \(\{x_i| i\in[1,8]\}\) and the correct option \(x_\alpha\), admits the following variational lower bound \cite{InfoNCE}:
\begin{align}\label{eq101}
I(x_\alpha; \{x_i\}_{i=1}^8) \geq -\mathbb{E}\left[ \log \frac{\frac{1}{K} \sum_{{\hat \alpha}\neq \alpha} P(x_{\hat \alpha}|\{x_i\}_{i=1}^8)}
{P(x_\alpha|\{x_i\}_{i=1}^8)} \right]
\end{align}
Here, \(P(x_\alpha | \{x_i\}_{i=1}^8)\) represents the true probability that \(x_\alpha\) is the correct option given the context matrix \(\{x_i| i \in [1,8]\}\), i.e., that \(x_\alpha\) can serve as the solution. \(K\) denotes the total number of potential incorrect options.

If we consider \( \hat P(x_\alpha | \{x_i\}_{i=1}^8, \theta) \) as a reliable approximation of the true conditional distribution \( P(x_\alpha | \{x_i\}_{i=1}^8) \), then substituting Equation (\ref{eq10}) into Equation (\ref{eq101}) yields a mutual information lower bound parameterized by \( \text{DIO}\):
\begin{align}\label{eq11}
I(x_\alpha; \{x_i\}_{i=1}^8) \geq -\mathbb{E}\left[ \log \frac{\frac{1}{7}\sum_{\substack{{\hat \alpha} = 9\\{\hat \alpha}\neq \alpha}}^{16} e^{\text{DIO}(\{x_i, x_{\hat \alpha}|i\in[1,8]\})}}
{e^{\text{DIO}(\{x_i, x_\alpha|i\in[1,8]\})}} \right]
\end{align}
It can be observed that the parameterized lower bound in Equation~\eqref{eq11} matches DIO's optimization objective in Equation~\eqref{eq102}:
\begin{align}\label{eq13}
    -\mathbb{E}\left[ \log \frac{\frac{1}{7}\sum_{\substack{{\hat \alpha} = 9\\{\hat \alpha}\neq \alpha}}^{16} e^{\text{DIO}(\{x_i, x_{\hat \alpha}|i\in[1,8]\})}}
{e^{\text{DIO}(\{x_i, x_\alpha|i\in[1,8]\})}} \right] = \log 7-\ell_\text{DIO}
\end{align}
Equation (\ref{eq13}) reveals that DIO's learning objective (i.e., maximizing $-\ell_\text{DIO}$) is neither human-like reasoning logic embedded in RPM problems nor even directly optimizing mutual information $I(x_\alpha; \{x_i\}_{i=1}^8)$, but rather maximizing a variational lower bound on that mutual information. 

Furthermore, by the properties of logarithmic functions, the following inequality holds:
\begin{align}\label{loss_bound}
-\log \hat P(x_\alpha|\{x_i\}_{i=1}^8,\theta)>{\ell}_\text{DIO}  
\end{align}
The left-hand side of Inequality \eqref{loss_bound} is precisely the expression for the loss function when DIO is optimized using the cross-entropy (CE) loss.
Notably, inequality \eqref{loss_bound} abstracts away the specific feed-forward architecture of DIO, primarily retaining its inductive mechanism. This implies that the same inequality holds for any model parameterizing the distribution $\hat{P}(x|\{x_i\}_{i=1}^8,\theta)$.
Consequently, it can be inferred that for those prior end-to-end RPM-solving models trained with the CE loss function, their learning objectives amount to optimizing a looser lower bound on $I(x_\alpha;\{x_i\}_{i=1}^8)$.

Equation (\ref{eq13}) and Inequality~\eqref{loss_bound} jointly uncover a pivotal fact: whether training the DIO model using \(\ell_{\text{DIO}}\) or previous end-to-end RPM solving models via the CE loss, their learning objectives are essentially maximizing lower bounds on the mutual information \(I(x_\alpha;\{x_i\}_{i=1}^8)\). As a result, regardless of how sophisticated the model architectures are, the tightness of these bounds directly and significantly determines the performance of such models. This observation further implies that in RPM problems the model architecture optimization is intrinsically limited; the crux is directing the model's objective to maximize a tighter lower bound on the mutual information.

\subsection{Tightening the Lower Bound via Hypothetical Options}

This paper contends that if DIO parameterizes a tighter variational lower bound and its learning objective explicitly targets this bound, we can more effectively maximize the mutual information \(I(x_{\alpha}; \{x_{i}\}_{i = 1}^{8})\), thereby further enhancing DIO’s performance.
By analyzing Equation (\ref{eq101})--\eqref{eq13}, we deduce that increasing the number \(K\) of incorrect options presented to DIO can steer its learning objective toward a tighter variational lower bound. However, each RPM instance provides only seven such options. Given this, we propose supplying DIO with additional hypothetical incorrect options.

Specifically, we introduce a set of hypothetical options, denoted as $\{x_i | i \in [17,\beta]\}$, into the original RPM instance $\{x_i | i \in [1,16]\}$, allowing DIO to recast the parameterized distribution $\hat{P}(x_\alpha | \{x_i\}_{i=1}^8, \theta)$ from the form given in Equation~\eqref{eq10} into the following:
\begin{align}\label{eq14}
    \hat P(x_\alpha|\{x_i\}_{i=1}^8,\theta)= \frac{e^{\text{DIO}(\{x_i,x_\alpha|i\in[1,8]\})}}{\sum_{{\hat \alpha} = 9}^{\beta}e^{\text{DIO}(\{x_i,x_{\hat \alpha}|i\in[1,8]\})}}
\end{align}
Correspondingly, the variational lower bound on mutual information $I(x_\alpha; \{x_i\}_{i=1}^8)$ is updated as follows:
\begin{align}\label{eq15}
    I(x_\alpha; \{x_i\}_{i=1}^8)\geq -\mathbb{E}\left[ \log \frac{\sum_{\substack{{\hat \alpha} = 9\\{\hat \alpha}\neq \alpha}}^{\beta} e^{\text{DIO}(\{x_i, x_{\hat \alpha}|i\in[1,8]\})}}
    {(\beta-9)\cdot e^{\text{DIO}(\{x_i, x_\alpha|i\in[1,8]\})}} \right]
\end{align}
Intuitively, this new lower bound is tighter. After incorporating the safeguard $\delta$, adopting it as the loss function for DIO should lead to improved performance.

However, this remains merely superficial. As shown in Equation (\ref{eq15}), if these introduced hypothetical options fail to influence DIO’s decisions, this new lower bound, when used as an optimization objective, will not bring DIO any significant improvement.
Specifically, when the probabilities that DIO assigns to the hypothetical options $\{x_i | i \in [17,\beta]\}$ approach zero, whereby these options cease to perturb the original distribution $\hat{P}(x_{{\alpha}} | \{x_i\}_{i=1}^{8}, \theta)$, the variational lower bounds derived from Equation (\ref{eq11}) and (\ref{eq15}) become equivalent when employed as loss functions.
This demonstrates that unconstructive options, which are readily assigned low weights and low probabilities by DIO, contribute nothing to tightening the lower bound.

In a nutshell, this paper proposes introducing extra constructive hypothetical options into DIO, such options that can reinforce acceptance of the correct options and strengthen repulsion of the incorrect ones, so as to steer its learning objective toward a tighter variational lower bound on mutual information \(I(x_{\alpha}; \{x_{i}\}_{i = 1}^{8})\) and, in turn, ultimately further enhance its performance.

\section{Brando: Tightening the Variational Lower Bound on Mutual Information by Constructing Hypothetical Options.}

This paper proposes incorporating additional constructive hypothetical options into DIO's training process as a promising avenue for enhancing its performance. To this end, this section introduces a training framework, Brando (Bounded Random Options Mapping Method).

\subsection{Trainable Hypothetical Opions.}

The RPM problem expects the solving model to be trained with only limited supervision, such as the index of the correct option. Relying on manual design to create constructive options contradicts this intent; we therefore propose learning a set of constructive options instead.

We establish a set of trainable hypothetical options $\{x_i | i \in [17, \beta]\}$ for the context matrix $\{x_i| i\in[1,8]\}$. This paper is not concerned with the specific attributes of these options; rather, it focuses solely on how much these options can alter DIO's evaluation of the original options $\{x_i | i \in [9, 16]\}$. In other words, we are solely concerned with the constructiveness of these hypothetical options with respect to the context matrix $\{x_i| i\in[1,8]\}$ in DIO, rather than their specific configurations. 
To this end, we design a new loss function $\ell_\text{Brando}$ to train these hypothetical options for constructiveness:
\begin{align}\label{Brando_loss}
    \ell_\text{Brando}=\ell_\text{up}+\frac{\ell_{down}}{\beta-17}
\end{align}
 where the loss term $\ell_\text{up}$:
 \begin{align}\label{ell_up}
    \ell_\text{up}=\log(\frac{\sum_{\substack{{\hat \alpha} = 9\\{\hat \alpha}\neq \alpha}}^{\beta}e^{\text{DIO}(\{x_i,x_{\hat \alpha}|i\in[1,8]\})}}
    {e^{\text{DIO}(\{x_i,x_\alpha|i\in[1,8]\})}}+\delta)- \log \delta
\end{align}
and the loss term $\ell_\text{down}$:
\begin{align}\label{ell_down}
    \ell_\text{down}=\sum_{{\hat\beta}=17}^{\beta}\log(\frac{\sum_{\substack{{\hat \alpha} = 9\\{\hat \alpha}\neq \alpha}}^{16}e^{\text{DIO}(\{x_i,x_{\hat \alpha}|i\in[1,8]\})}}{e^{\text{DIO}(\{x_i,x_{\hat\beta}|i\in[1,8]\})}}+\delta)- \log \delta
\end{align}
In this novel loss function:
\begin{enumerate}
    \item The loss term $\ell_\text{up}$ requires DIO to be able to identify the correct option from a pool that combines the original candidate options $\{x_i | i \in [9, 16]\}$ with the hypothetical options $\{x_i | i \in [17, \beta]\}$. It serves the purpose of tightening the variational lower bound on mutual information.

    \item The loss term $\ell_\text{down}$ encourages hypothetical options $\{x_i | i \in [17, \beta]\}$ to attain higher probabilities  than the original incorrect options $\{x_i | i \in [9, 16],i\neq \alpha\}$ under DIO. It acts as a safeguard to prevent hypothetical options from degenerating into unconstructive ones.
\end{enumerate}
By adopting $\ell_\text{Brando}$ as the training loss function, DIO not only obtains constructive hypothetical options but also parameterizes a tighter variational lower bound on $I(x_\alpha; \{x_i\}_{i=1}^8)$.

\subsection{Map Out Hypothetical Options}

Formulas~\eqref{ell_up} and~\eqref{ell_down} indicate that hypothetical options should demonstrate constructiveness for any given context matrix, meaning that such options should be customized according to the provided context matrix. This implies that naively training a set of shared hypothetical options across all context matrices is suboptimal. Given this, we require a mapping with the following form:
\begin{align}
    \{x_i|i\in [17,\beta]\}=\mathcal{F}(\{x_i|i\in [1,8]\})
\end{align}
Clearly, the mapping \(\mathcal{F}(\cdot)\) needs to yield structurally diverse and numerically flexible (\(\beta-17\)) hypothetical options conditioned on the context matrix \(\{x_i | i \in [1,8]\}\), implying an inherent many-to-many property. This indicates that conventional deep neural networks struggle to fulfill the role of mapping $\mathcal{F}(\cdot)$. Firstly, constructing a network with variable and dynamic output specifications is highly challenging. Moreover, from the perspective of neural network optimization, achieving many-to-many mappings in a naive end-to-end fashion is infeasible. Given these challenges, we propose a many-to-one mapping as a substitute:
\begin{align}
x_{\hat{i}}=\hat{\mathcal{F}}(\{x_{i},o_{\hat{i}}|i\in [1,8]\}), \,o_{\hat{i}}\sim \mathcal N(\mu,\sigma^2)
\end{align}
The task of the new mapping $\hat{\mathcal{F}}(\cdot)$ is to transform a Gaussian sample $o_{\hat{i}}$ into a constructive hypothetical option $x_{\hat{i}}$ tailored to the context matrix $\{x_i | i \in [1,8]\}$, a role that conventional networks can readily undertake.
In this setting, if we consider repeatedly drawing random seeds $o_{\hat{i}}$ from $\mathcal N(\mu,\sigma^2)$ and invoking $\hat{\mathcal{F}}(\cdot)$ each time, then the number of drawn seeds directly determines the number of derived options:
\begin{align}
&\{x_{\hat{i}}|i\in [17,\beta]\}=\hat{\mathcal{F}}(\{x_{i},o_{\hat{i}}|i\in [1,8]\})\mid_{\hat{i}=17}^\beta
\end{align}
Furthermore, provided $\hat{\mathcal{F}}(\cdot)$ avoids mode collapse, the diversity inherent in the drawn random seeds $\{o_{\hat{i}}|i\in [17,\beta]\}$ has the potential to be translated into the diversity of the derived hypothetical options $\{x_{\hat{i}}|i\in [17,\beta]\}$ by this mapping.
These two properties show that mapping \(\hat{\mathcal{F}}(\cdot)\), while operating as a many-to-one function, supports variable output specifications and furnishes diverse hypothetical options, thereby achieving the practical effect of a many-to-many mapping and enabling it to supplant the original mapping \(\mathcal{F}(\cdot)\).

\subsection{Instantiate Mapping \(\hat{\mathcal{F}}\) As a Network.}

\begin{figure}[htp]\centering
	\includegraphics[trim=0cm 0cm 0cm 0cm, clip, width=8.5
 cm]{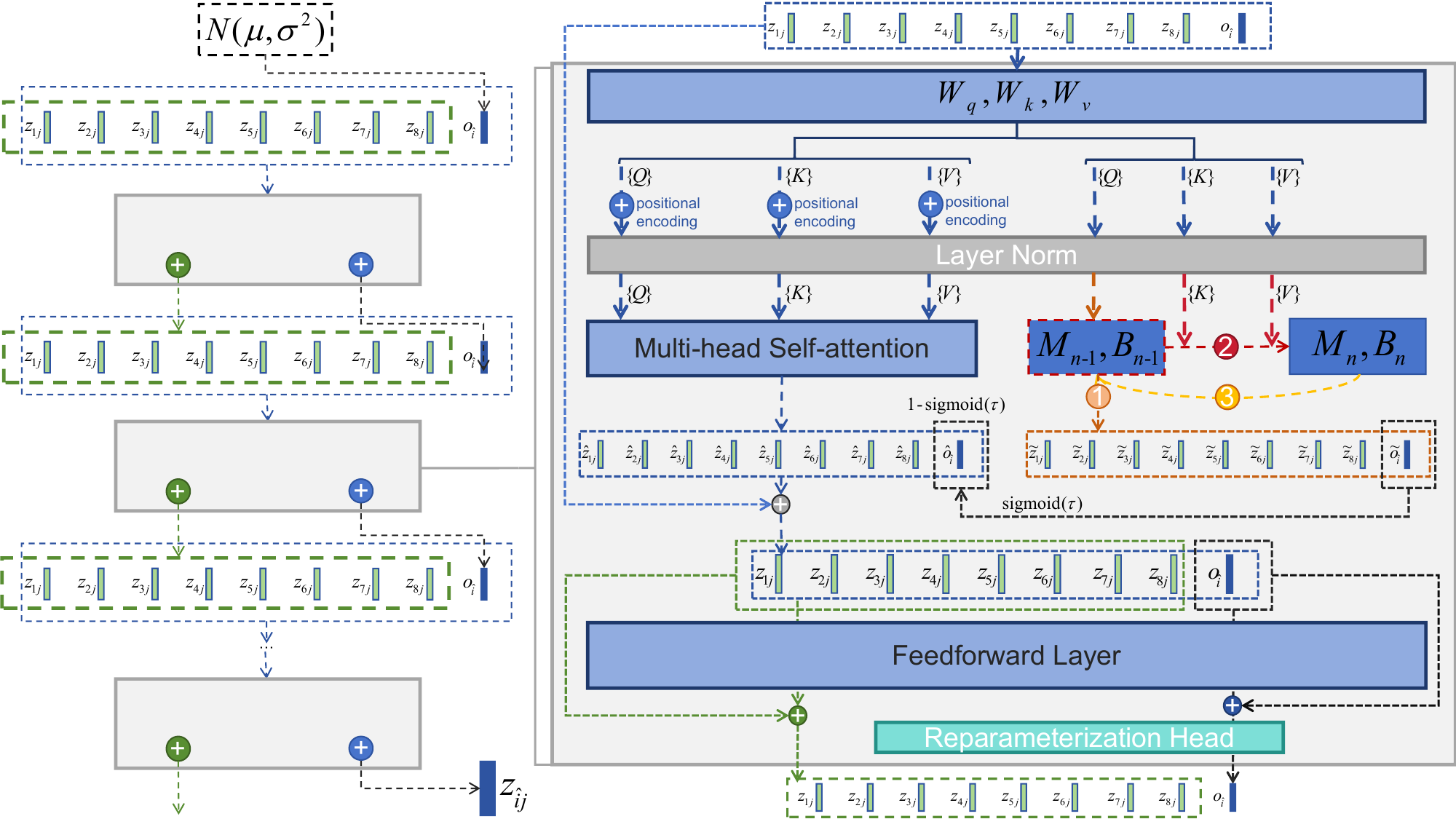}
	\caption{The Brando network.}
\label{Brando network}
\end{figure}

Given that mapping $\hat{\mathcal{F}}(\cdot)$ is both effective and feasible, we design a network named Brando to implement this new mapping:
\begin{align}\label{Brando}
z_{\hat{i}j}=\text{Brando}(\{z_{ij},o_{\hat{i}}|i\in [1,8]\}),\, o_{\hat{i}}\sim \mathcal N(\mu,\sigma^2)
\end{align}
Formula \eqref{Brando} shows that we abandon the option-mapping paradigm in favor of a representation-mapping approach, in order to reduce the computational load on Brando.
The structure of the Brando network is shown in the Figure \ref{Brando network}.

As indicated in Figure \ref{Brando network}, the Brando network is formed by stacking multiple logical layers, with no parameter sharing among each layer. The input to each logical layer consists of $\{z_{ij} | i \in [1,8]\}$ and $o_{\hat{i}}$. Except for the first layer, the input to all subsequent layers comes from the output of the preceding layer. 
For the first layer, $\{z_{ij} | i \in [1,8]\}$ are the features extracted by DIO from the context matrix, while the $o_{\hat{i}}$ is obtained through the reparameterization trick \cite{VAE} from a learnable Gaussian distribution $\mathcal N(\mu,\sigma^2)$ parameterized by mean $\mu$ and variance $\sigma^2$.
For the final layer, this paper treats its output $o_{\hat{i}}$ as the representation $z_{\hat{i}j}$ of a hypothetical option $x_{\hat{i}}$ provided by Brando with respect to the context matrix $\{x_i|i\in[1,8]\}$.

For the specific structure of each logical layer, it consists of a set of mapping matrices $\{W_q, W_k, W_v\}$, a multi-head self-attention mechanism, a feedforward layer (FFL), a dynamic memory bank $M_n$, a global buffer $B_n$, a learnable scalar $\tau$, and a reparameterization head. The feedforward process of each logical layer is as follows:
\begin{enumerate}
    \item When the features $\{z_{ij}, \hat{o}_{\hat{i}} | i \in [1,8]\}$ are input into a single logical layer, the mapping matrices $W_q$, $W_k$, and $W_v$ transform them into query ($Q$), key ($K$), and value ($V$) representations, respectively.

    \item After these $Q$, $K$, and $V$ representations are appended with positional encoding and processed by layer normalization, the multi-head self-attention mechanism computes interactions among them, yielding the attention outputs $\{\hat{z}_{ij}, \hat{o}_{\hat{i}} | i \in [1,8]\}$:

    \item As indicated by computation step 1 in Figure \ref{Brando network}, the normalized $Q$ representations employ a linear attention mechanism to learn the historical input feature information stored in the memory bank $M_{n-1}$ and buffer $B_{n-1}$, resulting in the memory outputs $\{\tilde{z}_{ij}, \tilde{o}_{\hat{i}} | i \in [1,8]\}$. The learning process can be interpreted as:
    \begin{align}
    \{\tilde{z}_{ij}, \tilde{o}_{\hat{i}} | i \in [1,8]\}  = \frac{\phi(Q)M_{n-1}}{\phi(Q)B_{n-1}}
    \end{align}
    where the basis function $\phi(\cdot)$ is an element-wise ELU+1 operation.

    \item As indicated by computation steps 2 and 3 in Figure \ref{Brando network}, the memory bank \( M_{n-1} \) and buffer \( B_{n-1} \) are updated by leveraging the \( K \) and \( V \) representations derived from the current input features, an update process inspired by infinite attention \cite{infin_atten} and expressed as:
    \begin{align}
    M_n &= M_{n-1} + \phi(K)^T\left(V - \frac{\phi(K)M_{n-1}}{\phi(K)B_{n-1}}\right) \label{M}\\
    B_n &= B_{n-1} + \sum \phi(K)\label{B}
    \end{align}

    \item After the learnable scalar $\tau$ is compressed into a scaling coefficient via the sigmoid function, a weighted summation is performed on the memory output $\tilde{o}_{\hat{i}}$ and attention output $\hat{o}_{\hat{i}}$ using this coefficient. The summation result is utilized to update $\hat{o}_{\hat{i}}$, enabling it to balance current reasoning with historical feature information. The update process of $\hat{o}_{\hat{i}}$ can be expressed as:
    \begin{align}\label{infinite}
    \hat{o}_{\hat{i}}\leftarrow[1-\text{Sigmoid}(\tau)]\hat{o}_{\hat{i}}+\text{Sigmoid}(\tau)\tilde{o}_{\hat{i}}
    \end{align}

    \item The attention outputs $\{\hat{z}_{ij}, \hat{o}_{\hat{i}} | i \in [1,8]\}$ (where $\hat{o}_{\hat{i}}$ has been updated), after undergoing residual connection with the input features, are processed by the feedforward layer and then subjected to another residual connection with their unprocessed state, ultimately yielding the logical outputs $\{z_{ij}, o_{\hat{i}} | i \in [1,8]\}$. These two residual connections can be formalized as:
    \begin{align}
    &Z\leftarrow \{{z}_{ij} , {o}_{\hat{i}}| i \in [1,8]\}+\{\hat{z}_{ij} , \hat{o}_{\hat{i}}| i \in [1,8]\}\\
    &\{{z}_{ij}, {o}_{\hat{i}} | i \in [1,8]\}\leftarrow\text{FFL}(Z)+Z
    \end{align}

    \item Importantly, the component ${o}_{\hat{i}}$ in the logical output $\{z_{ij}, o_{\hat{i}}| i \in [1,8]\}$ must be processed by the reparameterization head before serving as the final output. The reparameterization head operates as follows:
    \begin{align}
    & o_{\mu}, o_{\ln\sigma}=\text{MLP}({o}_{\hat{i}})\\
    &{o}_{\hat{i}}\leftarrow {o}_{\mu}+e^\frac{o_{\ln\sigma}}{2}\cdot \epsilon
    \end{align}
    The $\epsilon$ represents a random sample drawn from $\mathcal N(0,1)$ and MLP$(\cdot)$ refers to a Multi-Layer Perceptron capable of doubling the dimension of ${o}_{\hat{i}}$.
\end{enumerate}
As indicated in Formulas~\eqref{M}-\eqref{B}, in each logical layer, the memory bank $M_n$ and the buffer $B_n$ are dynamically updated as each new batch of context matrices is input. Therefore, their initial values must be explicitly defined:
\begin{align}
M_0=0,\, B_0=1\times 10^{-9}
\end{align}
Furthermore, the first update of the memory bank and the buffer can be explained as follows:
\begin{align}
M_1 = \phi(K)^TV,\,B_1 = \sum \phi(K) + 10^{-9}
\end{align}

From the learnable Gaussian distribution ${N}({\mu}, {\sigma}^2)$, we sample random vectors ${o}_{\hat{i}}$, which undergo sequential refinement in the {Brando} network through dynamic interaction with both current context matrix representations and historical image representations accumulated over multiple rounds of training iterations. This iterative conditioning process enables ${o}_{\hat{i}}$ to progressively assimilate visual patterns observed throughout the training trajectory, ultimately yielding the ultimate hypothesis representation ${z}_{\hat{i}j}$. Moreover, alterations in the initial promoter ${o}_{\hat{i}}$ are capable of inducing changes in the ultimate option representation $z_{\hat{i}j}$. This satisfies the requirement to map diverse hypothetical options for the context matrix.
 
\subsection{The Combination of Brando and DIO.}

\begin{figure}[htp]\centering
	\includegraphics[trim=0cm 0cm 0cm 0cm, clip, width=8.5
 cm]{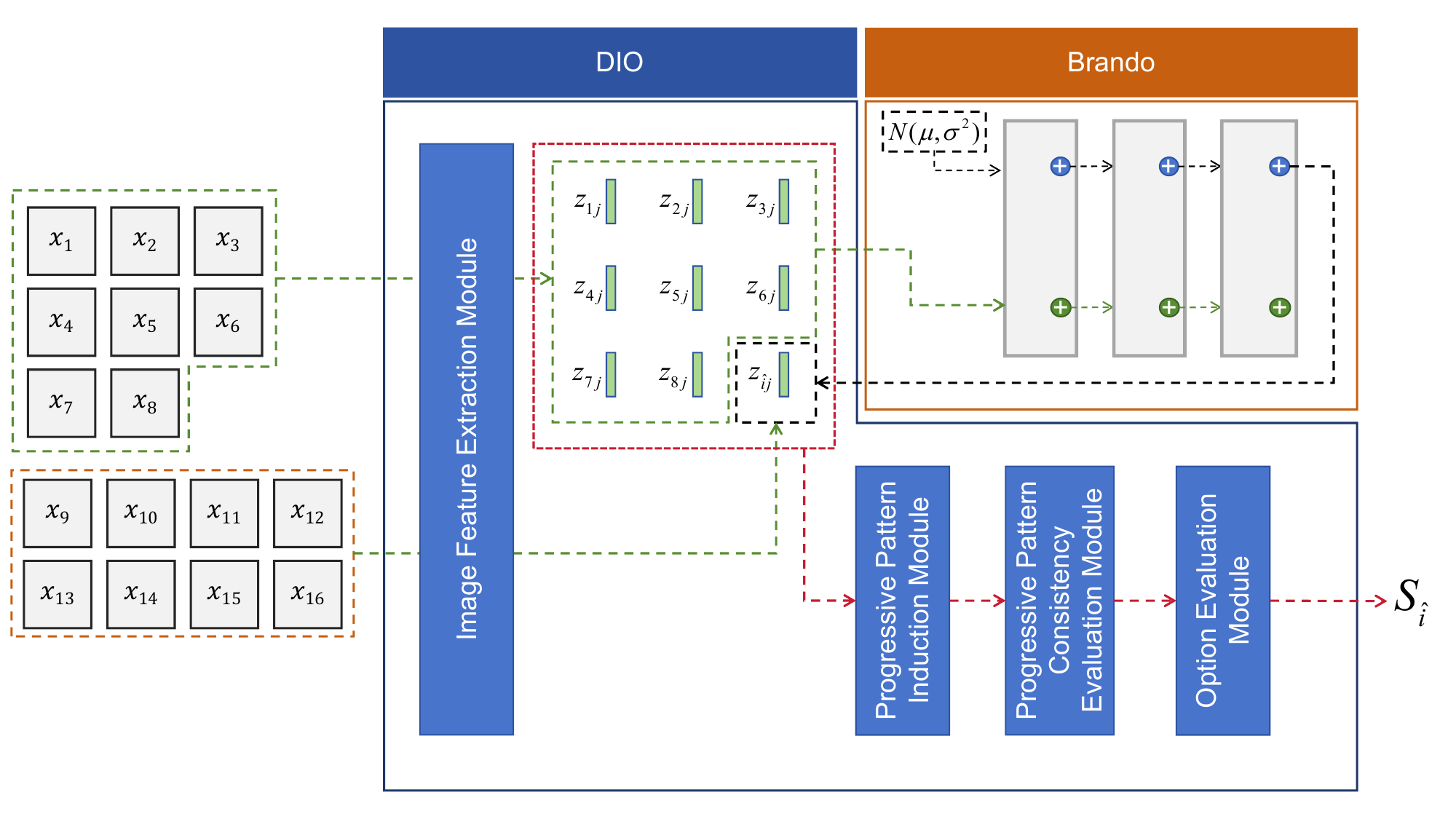}
	\caption{The Combination Method of DIO and Brando.}
\label{The Combination Method of DIO and Brando}
\end{figure}

Prior to integrating the Brando network with DIO, DIO must be pre-trained on the original option pool $\{x_i| i \in [9,16]\}$ using the loss function 
$\ell_{\text{DIO}}$. Once DIO is pre-trained to reach a state close to its optimal performance, the Brando network will be integrated with DIO according to the method illustrated in Figure \ref{The Combination Method of DIO and Brando}.
Subsequently, the loss function will be switched from $\ell_{\text{DIO}}$ to $\ell_{\text{Brando}}$ to continue training this integrated system, thereby further optimizing DIO's performance.
Another key point is that during the joint training process, DIO's parameters excluding those in the image feature extraction module will be frozen every other iteration.

\section{WORLD: Tightening the Variational Lower Bound on Mutual Information Through Deconstructing the Problem Space}

It is readily apparent that the architecture of the Brando network leaves substantial room for theoretical refinement. The pursuit of a more optimal structure for implementing the mapping $\mathcal{F}(\cdot)$ appears to be an open-ended endeavor, with no discernible asymptote in sight. Moreover, providing a sufficient number of hypothetical options for DIO using Brando entails considerable computational overhead. Therefore, this paper argues that Brando is not a flawless method. 
To this end, this paper proposes a method named WORLD (Weighted Option-Represented Lower-bound Delimitation Method), which breaks away from the paradigm of the mapping $\mathcal{F}(\cdot)$. This method can also tighten the variational lower bound of mutual information $I(x_{\alpha}; \{x_{i}\}_{i = 1}^{8})$.

\subsection{Local Estimation $<$ Global Estimation $<$ Global Modeling.}
Formula (\ref{eq11}) indicates that the key to guiding the learning objective of solving model towards a tighter lower bound of the mutual information $I(\{x_i\}_{i=1}^8;x_\alpha)$ lies in providing additional incorrect options to the model for each context matrix. Moreover, this paper also points out that the introduced additional incorrect options must avoid being unconstructive.

The Brando network essentially functions as a distribution transformation mechanism, whose task is to map samples \(o_{\hat{i}}\) drawn from a Gaussian distribution \(\mathcal N(\mu, \sigma^{2})\) into features \(z_{\hat{i}j}\) of incorrect options customized for a given context matrix. This paper argues that instead of implicitly estimating the distribution of incorrect option features for each given context matrix, it is more effective to explicitly estimate the distribution of all RPM image features \(z_{ij}\). In this way, when confronted with a specific context matrix, the required incorrect option features can be sampled in a targeted manner from this estimated distribution.

However, accurately estimating the distribution of RPM image features $ z_{ij} $ poses significant challenges, and even if such a distribution were successfully estimated, targeted sampling from it would remain highly non-trivial. Therefore, this paper argues that explicitly modeling the distribution of $ z_{ij} $ during the process where the DIO's feature extraction module encodes $ x_i $ into $ z_{ij} $ constitutes an effective approach.
Because this method can obtain the distribution of $z_{ij}$ without estimation and facilitate targeted sampling by directly controlling the distribution.

This paper has found that the Gaussian Mixture Model (GMM) \cite{GMM} offers remarkable advantages, as it not only facilitates efficient modeling but also enables targeted sampling. Suppose the features \(z_{ij}\) extracted by DIO from RPM images follow a GMM with known numbers of components and parameters. Then, for a given context matrix, after identifying the components corresponding to its correct options within this GMM, we can reasonably regard the components not corresponding to the correct options as the source of incorrect option features. Based on this, we are able to sample comprehensive and diverse incorrect option features from these specific components for DIO to tighten its learning objective. 
Given these advantages, we shift our focus to transforming the underlying distribution of RPM image features into a GMM.

\subsection{Model the Distribution of RPM Image Features as a GMM.}

To model the distribution of features encoded by DIO for RPM images as a GMM, this paper proposes the WORLD method.
Specifically, this paper introduces a trainable vector set $\{U_m | m \in [1, M]\}$ and considers it as the mean vectors of a GMM. Using this approach, we define a GMM comprising $M$ components, where the means of all components are trainable while the variances remain unknown. Subsequently, by projecting the image features $z_{ij}$ encoded by DIO onto this constructed GMM framework, we can model the distributional characteristics of $z_{ij}$.

\begin{figure}[htp]\centering
	\includegraphics[trim=0cm 4cm 0cm 2.4cm, clip, width=8.5
 cm]{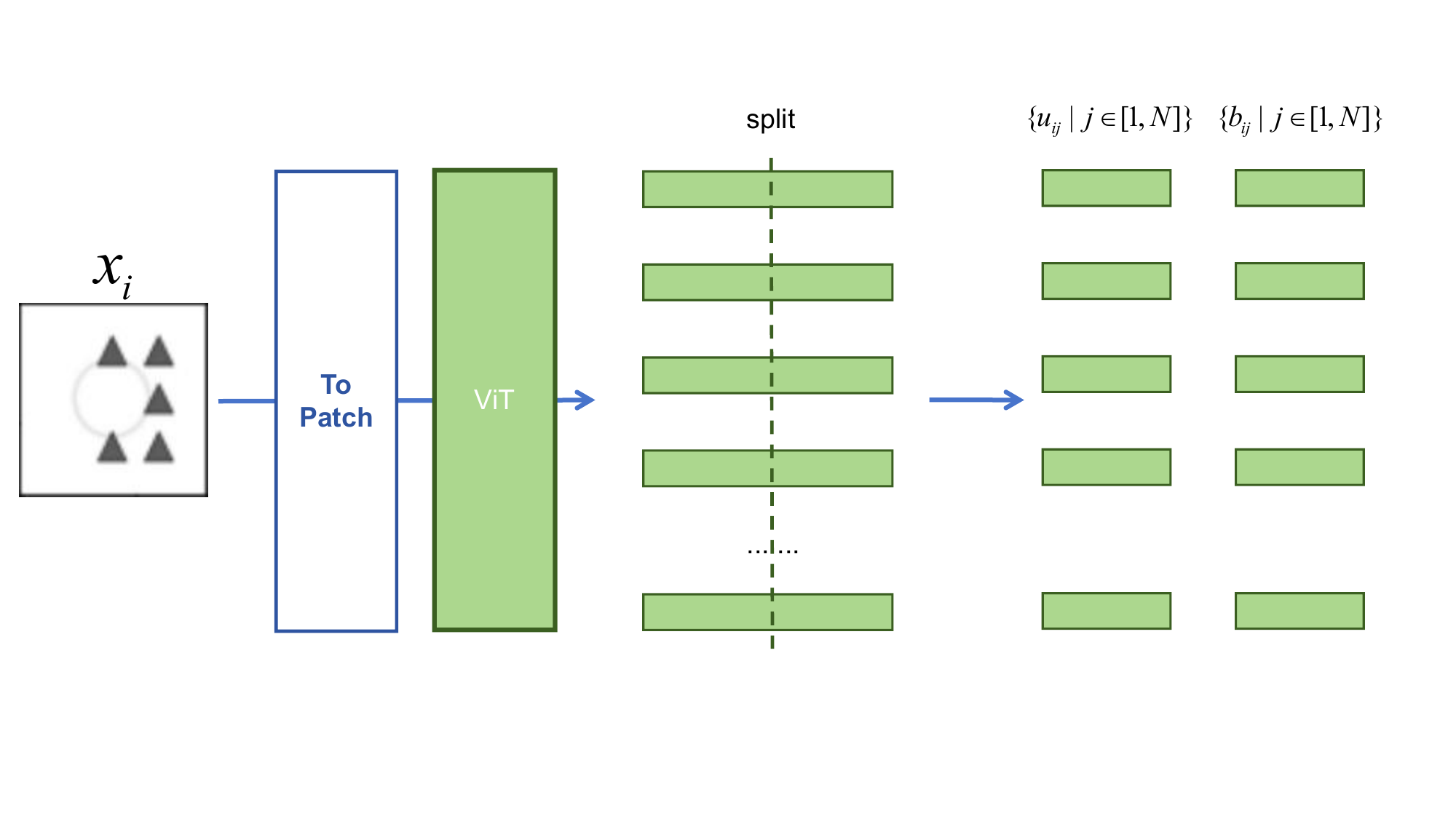}
	\caption{The modified Image Feature Extraction Module.}
\label{The modified Image Feature Extraction Module.}
\end{figure}

In order to enable the features $z_{ij}$ to be projected more easily onto the constructed GMM $\{U_m | m \in [1, M]\}$, the way DIO encodes features for RPM images requires minor modifications. The modifications are shown in Figure \ref{The modified Image Feature Extraction Module.}.
Specifically, doubling the overall width of the ViT within DIO's image feature extraction module can achieve the desired effect, causing the dimensions of the latent features encoded for RPM images to double accordingly. These doubled features are evenly split into two parts, denoted as \(u_{ij}\) and \(b_{ij}\). Then, their sum is taken as the new feature representation \(z_{ij}\) for the image \(x_i\). This updated module can be expressed by the following formula:
\begin{align}
    \{u_{ij}, b_{ij}\} = \text{WORLD-FEM}(x_i), \,z_{ij}=u_{ij}+b_{ij}\label{ud_FEM}
\end{align}
The new derivation method for $z_{ij}$ ensures that, when $b_{ij}$ follows a standard Gaussian distribution $\mathcal{N}(0,1)$, the distribution of $z_{ij}$ explicitly adheres to a Gaussian distribution $\mathcal{N}(u_{ij},1)$.
Given this, we merely need to assign the latent features $u_{ij}$ of all observed RPM images onto specific components of the constructed GMM $\{U_m | m \in [1, M]\}$ and make the latent features $b_{ij}$ follow the distribution $\mathcal N(0,1)$, so as to effectively model the distribution of $z_{ij}$ as a GMM with trainable means and known variances.

The aforementioned method for modeling the distribution of $z_{ij}$ can be implemented by introducing five additional loss function terms:
\begin{enumerate}
    \item  A loss term $\ell_1$ aims to encourage the mean of the distribution of \(z_{ij}\) to approximate the constructed GMM $\{U_m | m \in [1, M]\}$, and its mathematical expression can be formulated as:
    \begin{align}\label{weight_1}
 &\ell_1 =  \frac{1}{16\cdot N}\sum_{i=1}^{16}\sum_{j=1}^{N}\| u_{ij} - \text{gs}[{U}_{\hat{m}}]\|_2^2\\
  &\text{where} \quad \hat{m} = \underset{m\in [1,M]}{\text{argmin}} \| {u}_{ij} - {U}_m \|_2^2 \nonumber
    \end{align}
    Here, $ \text{gs}[\cdot] $ denotes the gradient stopping operation.

    \item  A loss term $\ell_2$ is introduced to enforce $b_{ij}$ to follow a standard normal distribution, expressed as:
    \begin{align}
 &\ell_2 =  \frac{1}{16\cdot N}\sum_{i=1}^{16}\sum_{j=1}^{N}\| b_{ij} - \epsilon_{ij}]\|_2^2\\
  &\text{where} \quad\epsilon_{ij} \sim \mathcal N(0,1)\nonumber
    \end{align}
    
    \item  A loss term $\ell_3$, which is utilized to train the components within the constructed GMM $\{U_m | m \in [1, M]\}$, can be expressed as:
    \begin{align}\label{weight_3}
 &\ell_3 =  \frac{1}{16\cdot N}\sum_{i=1}^{16}\sum_{j=1}^{N}\| {U}_{\hat{m}} - \text{gs}[u_{ij}]\|_2^2\\
 &\text{where} \quad \hat{m} = \underset{m\in [1,M]}{\text{argmin}} \| {u}_{ij} - {U}_m \|_2^2\nonumber
    \end{align}

    \item  A mutual information loss term $\ell_4$, which facilitates the training of components within the constructed GMM $\{U_m | m \in [1, M]\}$, is expressed as:
    \begin{align}
        \ell_4 &= \frac{1}{16}\sum_{i=1}^{16} \| x_{i} - \hat{x}_i \|_2^2 \label{111} \\
        \text{where} \quad 
        &\hat{x}_i = \mathcal{D}\left( \left\{ \hat{u}_{ij} + b_{ij} \mid j \in [1,N] \right\} \right), \nonumber \\
        &\hat{u}_{ij} = U_{\hat{m}} - \text{sg}[u_{ij}] + u_{ij}, \nonumber \\
        &\hat{m} = \mathop{\mathrm{argmin}}_{m \in [1,M]} \| u_{ij} - U_m \|_2^2. \nonumber
    \end{align}
    Obviously, the loss term $\ell_4$ is formulated to maximize the mutual information between the components within $\{U_m | m \in [1, M]\}$ and the latent features $u_{ij}$, via an auxiliary reconstruction task facilitated by the decoder $\mathcal{D}(\cdot)$. The decoder $\mathcal{D}(\cdot)$ is structured as a depth-reversed counterpart of the ViT, transforming latent features into a reconstructed image. 
    The computation of $\hat{u}_{ij}$ ensures the existence of a gradient propagation pathway: $\hat{x}_i \rightarrow U_{\hat{m}} \rightarrow \hat{u}_{ij} \rightarrow u_{ij} \rightarrow x_i$, thereby guaranteeing the effectiveness of the loss term $\ell_4$.

     \item   A mode collapse prevention loss term $\ell_5$ for the latent features $u_{ij}$, defined as:
\begin{align}\label{generate_ability_d}
    \ell_5 &= \frac{1}{16}\sum_{i=1}^{16} \| x_{i} - \tilde{x}_i \|_2^2  \\
    \text{where} \quad 
    &\tilde{x}_i = \mathcal{D}\left( \left\{ \hat{u}_{ij} + \epsilon_{ij} \mid j \in [1,N] \right\} \right), \nonumber \\
    &\epsilon_{ij} \sim \mathcal N(0,1)\nonumber\\
    &\hat{u}_{ij} = U_{\hat{m}} - \text{sg}[u_{ij}] + u_{ij}, \nonumber \\
    &\hat{m} = \mathop{\mathrm{argmin}}_{m \in [1,M]} \| u_{ij} - U_m \|_2^2. \nonumber
\end{align}
    It is not hard to notice that during the construction of the loss term $\ell_4$, there exists a variable calculation pathway, which can also be referred to as an information pathway, namely: $x_i \rightarrow b_{ij} \rightarrow \hat{x}_i$. The presence of this pathway may potentially lead to mode collapse in the latent features $u_{ij}$, thereby making it impossible to model the distribution of $z_{ij}$ as a GMM. Therefore, the loss term $\ell_5$ comes into being, ensuring that the latent feature $u_{ij}$ cannot be bypassed during the reconstruction process of the RPM image $x_i$.
\end{enumerate}
After performing a weighted summation of these five loss terms, we obtain the complete loss function $\ell_{\text{WORLD}}$:
\begin{align}\label{WORLD_loss}
    \ell_{\text{WORLD}}=\ell_1+
                        \ell_2+
                        \ell_3+
                        \lambda_{0}\cdot\ell_4+
                        \lambda_{1}\cdot\ell_5
\end{align}
Here, $\lambda_{0 }$ and $\lambda_{1}$ are set to $5$ to balance the inductive difficulties between these terms.

\begin{figure}[htp]\centering
	\includegraphics[trim=0cm 0cm 0cm 0cm, clip, width=8.5
 cm]{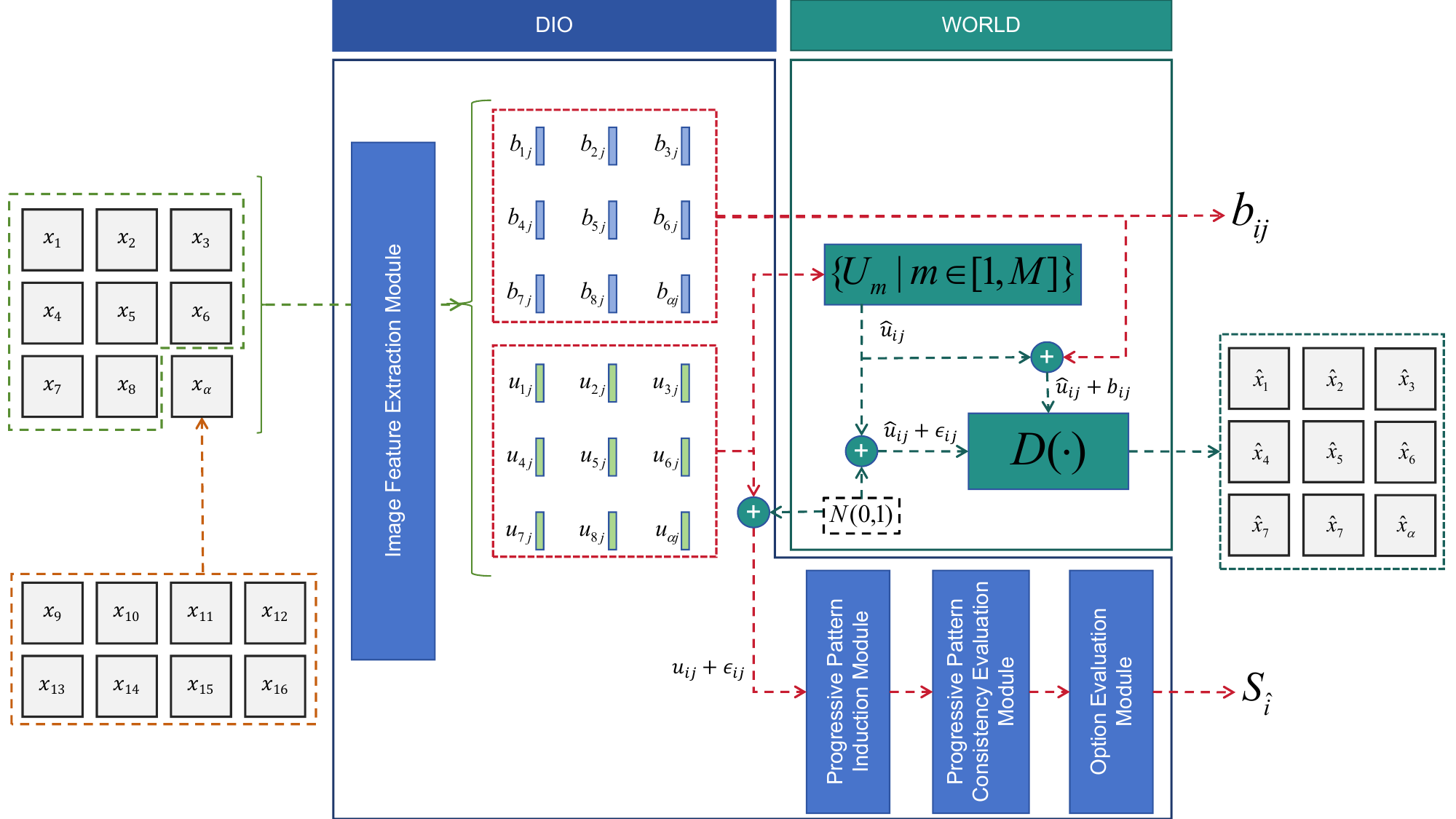}
	\caption{Feedforward Process of DIO Model Incorporating $\ell_\text{WORLD}$ as an Additional Loss Term.}
\label{Forward Feed Process of DIO Model Incorporating ℓ_WORLD as an Additional Loss Term}
\end{figure}

By employing $\ell_{\text{WORLD}} + \ell_{\text{DIO}}$ as the loss function, we can train a DIO model in which the distribution of image features $z_{ij}$ is modeled as a GMM. Notably, when training the DIO using $\ell_{\text{WORLD}} + \ell_{\text{DIO}}$, the feedforward mechanism undergoes the transformation illustrated in Figure \ref{Forward Feed Process of DIO Model Incorporating ℓ_WORLD as an Additional Loss Term}. 
As shown in Figure \ref{Forward Feed Process of DIO Model Incorporating ℓ_WORLD as an Additional Loss Term}, the latent variable fed into the progressive pattern induction module within DIO is no longer \( z_{ij} \), but rather \( u_{ij} + \epsilon_{ij} \). In other words, the feedforward process indicated by Equation (\ref{PPIM}) is updated as follows:
\begin{align}\label{PPIM_}
    \{P_{\alpha jkl}|j\in [1,N],k\in[1,9],l\in[1,4]\}\nonumber\\ 
    = \text{PPIM}(\{u_{ij}+\epsilon_{ij},u_{\alpha j}+\epsilon_{\alpha j}| i\in[1,8]\})|_{j=1}^N
\end{align}
Where $u_{ij} + \epsilon_{ij}$ can be regarded as a sample drawn from the GMM component $U_{\hat{m}}$ that feature $z_{ij}$ follows.
This design enables DIO to interact with a fully modeled GMM rather than deterministic image features, thereby training the DIO to make decisions based on the complete probabilistic structure of the GMM instead of specific feature instantiations.
 
\subsection{Targeted Sampling From the Learned GMM.}

To steer {DIO}'s learning objective towards a tighter variational lower bound of the mutual information $I(\{x_i\}_{i=1}^8;x_\alpha)$, we also need to sample sufficient and diverse incorrect option features from the modeled GMM $\{U_m | m \in [1, M]\}$ for each given context matrix. Thanks to the inherent properties of GMM, targeted sampling within the set $\{U_m | m \in [1, M]\}$ is relatively straightforward. For a given context matrix $\{x_i | i \in [1,8]\}$, after identifying the components in the set $\{U_m | m \in [1, M]\}$ that correspond to the provided correct option $x_\alpha$, samples randomly drawn from the remaining components can be considered, to a certain extent, as features of the incorrect options.

Identifying the components corresponding to the correct option, targeted sampling of the incorrect option features, and steering the learning objective of DIO can all be achieved by introducing an additional loss function term:
\begin{align}\label{bound}
    &{\ell_{\text{Weighted}}}\nonumber\\
    &= \sum_{j = 1}^{N}\log (\frac{\sum_{\substack{m = 1\\m\neq \tilde{m}}}^{M} {{e^{\frac{\tilde{\text{DIO}}(\{u_{ij} + \epsilon_{ij}, U_m + \epsilon_m | i \in [1,8]\})}{|\tau|}}}} }
    {{{e^{\frac{\tilde{\text{DIO}}(\{u_{ij} + \epsilon_{ij}, U_{\tilde{m}} + \epsilon_{\tilde{m}} | i \in [1,8]\})}{|\tau|}}}}}+\delta)+\log\delta\\
    &\text{where}\quad \tilde{m} = \underset{m\in [1,M]}{\text{argmin}} \| {u}_{\alpha j} - {U}_m \|_2^2,\,\,\epsilon_m\sim \mathcal N(0,1)\nonumber
\end{align}
$\tilde{\text{DIO}}$ refers to the DIO {without} the image feature extraction module, which is capable of processing image features into scores.
In RPM problems, any RPM image that conforms to the progressive patterns exhibited in the context matrices can be considered a correct option. This indicates that RPM problems inherently possess a multi-solution characteristic. Consequently, the rigid approach of designating samples drawn from components that do not correspond to the provided correct option $x_\alpha$ as features of incorrect options carries a potential risk of overgeneralization. 
To mitigate this sample classification bias, we introduce a trainable smoothing parameter $\tau$. 
$\tilde{m}$ identifies the index of the GMM component that best matches the feature $z_{\alpha j}$ (or $u_{\alpha j}$) of the correct option $x_{\alpha}$. 

The role of $\ell_{\text{Weighted}}$ is to fine-tune the DIO model. Specifically, when DIO approaches optimal performance under the constraint of the loss function $\ell_{\text{WORLD}} + \ell_{\text{DIO}}$, incorporating the additional loss term $\ell_{\text{Weighted}}$ to update the loss function to $\ell_{\text{WORLD}} + \ell_{\text{DIO}} + \ell_{\text{Weighted}}$ creates additional learning capacity and potential of performance improvement. This adjustment enables the introduction of {supplementary incorrect options} during DIO's late-stage training, thereby steering its learning objective toward a {tighter variational lower bound} on the mutual information $I(\{x_i\}_{i=1}^8; x_\alpha)$, ultimately leading to further performance enhancement.

\section{DIEGO: Methods for Rectifying the Causal Chain modeled by DIO}
This study has made two attempts to refine the learning objectives of the DIO model, resulting in the development of two distinct methodologies: Brando and WORLD. 
In this section, we devise a methodology named DIEGO (Direct Intervention for Enforcing Guided Optimal Reasoning), which is designed to rectify the causal chains modeled by DIO, thereby advancing its reasoning performance.

\subsection{Target Reasoning Rather Than Solving Problems}

This paper realizes that setting the maximization of the mutual information $I(x_\alpha; \{x_i\}_{i = 1}^8 )$ as the learning objective of DIO is suboptimal. Although this statistical measure strongly correlates with the solving accuracy of RPM tasks, it neglects the human-predefined reasoning logic \cite{Causal inference}, yielding a model optimized for high accuracy rather than faithful causal reasoning.
Thus, the performance gains brought by mutual information may be superficial.
Both the Brando and WORLD methodologies aim to assist DIO in maximizing mutual information, yet neither resolves this limitation.
In a nutshell, mutual-information-driven approaches are effective but need an explicit causal perspective.

Therefore, this paper argues that the causal chain underlying RPM problems should be further modeled. The network architecture of DIO is designed to mirror this causal chain, aiming for DIO to automatically capture the chain during training. 
Nevertheless, while this design guarantees structural alignment, it cannot ensure that the latent variables derived by DIO are also semantically aligned. Given that mutual information fails to account for causality \cite{Causal inference}, the semantic alignment awaits empirical verification.

\begin{figure}[htbp]\centering
	\includegraphics[trim=9cm 16cm 4.5cm 2.5cm, clip, width=8.5
 cm]{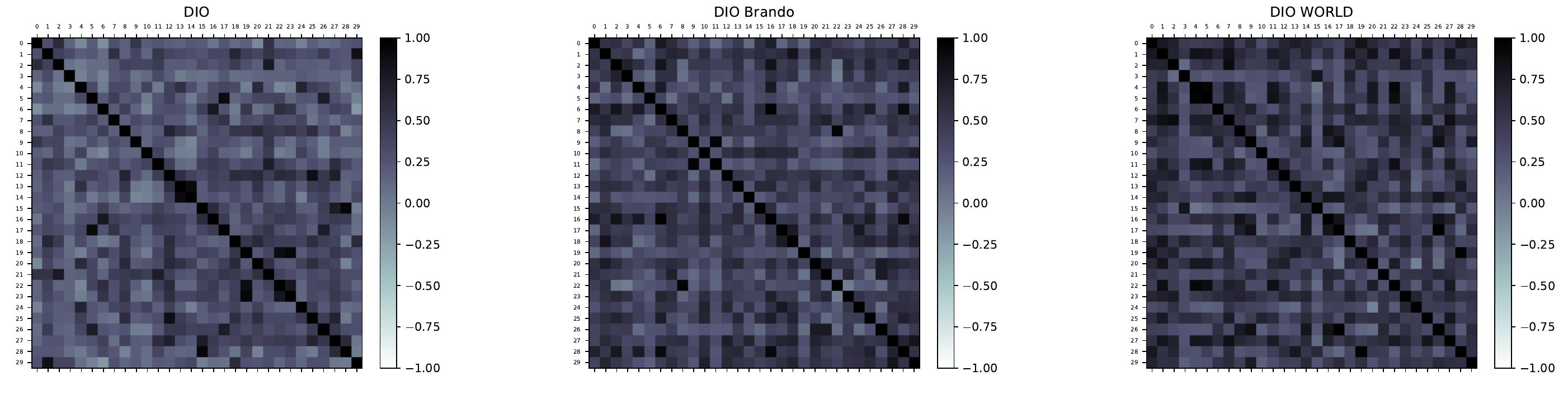}
	\caption{Heatmap of Progressive-Pattern Representation Extracted from the 3$\times$3 Grid under RAVEN Problems.}
\label{Heatmap}
\end{figure}

In order to verify semantic alignment, this paper takes the ``3$\times$3 Grid" sub-task from the RPM-style RAVEN benchmark \cite{RAVENdataset} as the research object. We conduct training on this dataset using DIO, as well as DIO supplemented with Brando and WORLD. Subsequently, we extract 30 instances that share strictly identical progressive patterns and compute the pairwise similarities among the progressive-pattern representations that DIO encodes for these 30 instances. 
The calculated $30\times 30$ similarity matrices are visualized as heatmaps in Figure \ref{Heatmap}.

If DIO accurately models the causal chain embedded in RAVEN problems, then once DIO encodes 30 instances sharing the same progressive patterns into pattern representations, those representations should exhibit high cosine similarity among themselves.
Contrary to expectations, the three similarity matrices in Figure \ref{Heatmap} fail to support this anticipated outcome. This finding indicates that, even with auxiliary supervision from Brando or WORLD, the causal chain learned by DIO systematically diverges from the ground-truth chain inherent in RPM problems. This further underscores the limitations of adopting mutual information $I(x_\alpha; \{x_i\}_{i = 1}^8 )$ as the learning objective. Therefore, we propose that further modeling of the causal chain could  prioritize the rectification of DIO’s semantic representations, which would substantially enhance its performance on RPM problems.

\subsection{Rectifying the Causal Chain Modeled by DIO.}

In preliminary experiments, we found that DIO exhibits deviations in modeling the segment of the causal chain that involves ``abstract attributes $\rightarrow$ progressive patterns of attributes". This corresponds to a misalignment between the outputs \( P_{\alpha{jkl}} \) produced by Formula~\eqref{PPIM} and human semantics. Therefore, this paper proposes  rectifying the progressive-pattern representations $P_{\alpha{jkl}}$ encoded by DIO as an effective remedy.

This paper notes that each RPM instance is accompanied by metadata describing its progressive patterns that precisely match the semantics predefined for $P_{\alpha{jkl}}$ in the DIO architecture.
We argue that incorporating this metadata as a supervisory signal into DIO's training process to guide the derivation of $P_{\alpha{jkl}}$ can rectify the causal chain modeled by DIO.

To this end, we propose the {DIEGO} method. 
DIEGO constructs a set of trainable reference vectors \(\{V_f | f \in [1,F]\}\). By assigning each vector a distinct metadata modality, this entire set can be considered a comprehensive metadata dictionary.
An auxiliary loss term \(\ell_{\text{DIEGO}}\) is introduced to align every progressive pattern representation \(P_{\alpha jkl}\) extracted by DIO with its corresponding reference vector in \(\{V_f \mid f \in [1,F]\}\) according to the accompanying metadata.  
This term is formulated as follows:
\begin{align}
    &{\ell _{\text{DIEGO}}}= -\sum_{\hat{f}=1}^{F} \sum_{j=1}^{N} \sum_{k=1}^{9}  \sum_{l = 1}^{2} \text{Meta}_{{\hat{f} l}}\cdot\log \frac{{{e^{\frac{\left({P_{\alpha jkl} }\cdot {V_{\hat{f}}}\right)}{|\tau|}}}}}{\sum_{f = 1}^{F} {{e^{\frac{\left({P_{\alpha jkl} }\cdot {V_{f}}\right)}{|\tau|}}}} }\label{act3}
\end{align}
Here, $\text{Meta}_{\hat{f} l}$ is a binary indicator that flags whether the $l$-th metadata entry matches the modality associated with the $\hat{f}$-th vector within the reference dictionary \(\{V_f | f \in [1,F]\}\). Meanwhile, $\tau$ is a trainable scaling coefficient. 
The detailed formulation of $\ell_{\text{DIEGO}}$ elucidates our alignment mechanism. Guided by the metadata-derived indicators, this loss term employs logarithmic smoothing and exponential scaling to regulate the cosine similarities between each $P_{\alpha_{jkl}}$ and all vectors in the reference dictionary. 

\begin{figure}[htp]\centering
	\includegraphics[trim=0cm 0cm 0cm 0cm, clip, width=8.5
 cm]{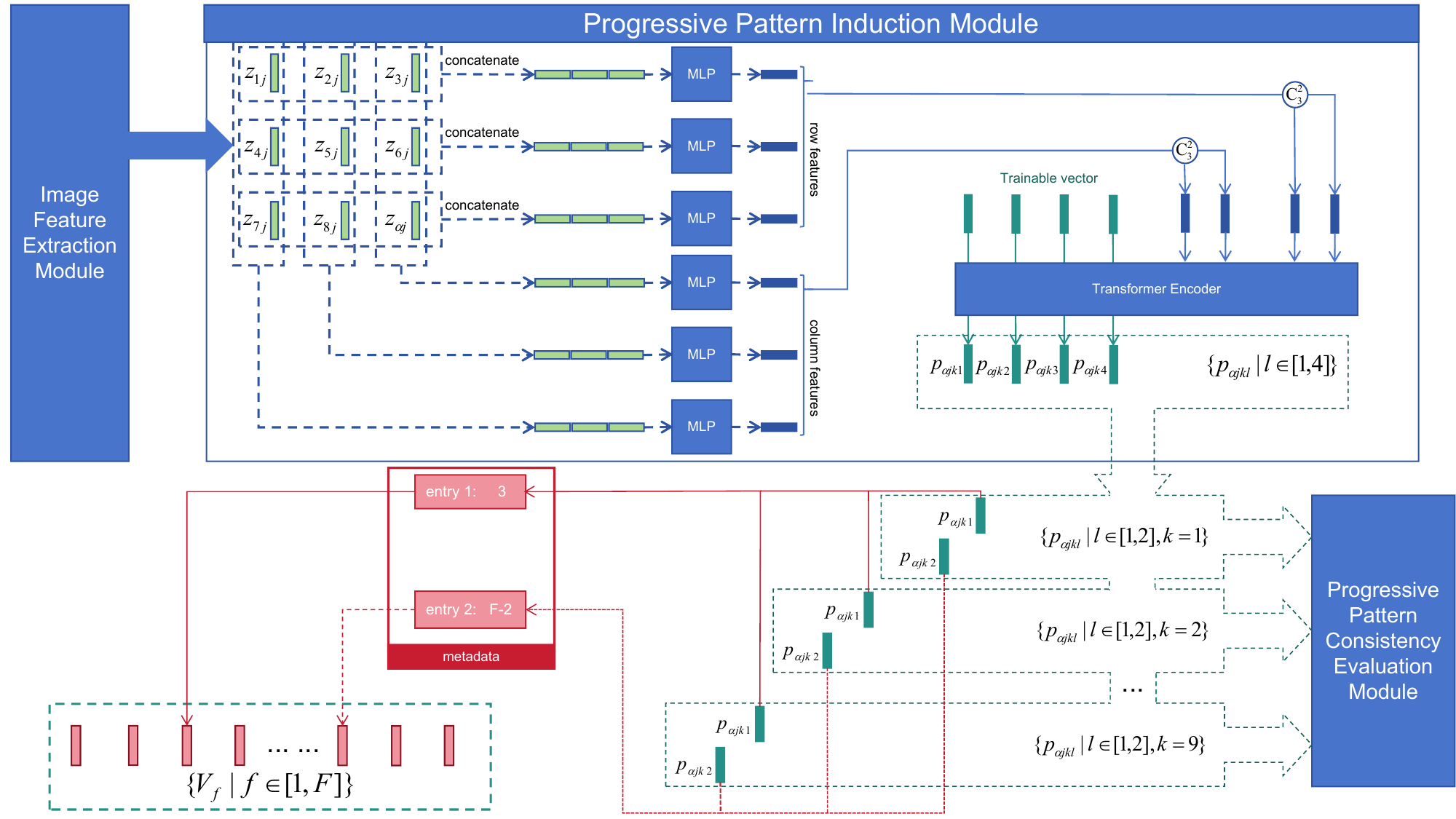}
	\caption{Feedforward Process of DIO Model Incorporating
$\ell_{\text{DIEGO}}$ as an Additional Loss Term.}
\label{Diego_fig}
\end{figure}

Evidently, our objective has shifted from statistical quantity control to vector similarity control, distinguishing DIEGO from Brando and WORLD in that it is no longer mutual-information-driven but progressive-pattern-driven.
When the DIO model is trained with the composite loss $\ell_{\text{DIO}} + \ell_{\text{DIEGO}}$ and the trainable reference dictionary $\{V_f |f \in [1,F]\}$, the forward pass of the relevant modules proceeds as illustrated  in Figure \ref{Diego_fig}.
After training, a solving model that captures causal chains in RPM problems is obtained.


\section{Solving Generative RPM Problems}

Generative RPM problems represent an advanced variant of RPM tasks. In these problems, the solving models are required to generate at least one solution that conforms to the progressive patterns within the given context matrix, rather than selecting from a predefined and bounded option pool, thereby more accurately reflecting the model’s comprehension of the reasoning logic. As solution identification remains progressive-pattern-driven, generative RPM problems inherit the multi-solution characteristic from conventional RPM tasks. Consequently, solving models are expected to produce multiple solutions that are structurally diverse, which poses significant challenges for most traditional neural networks. However, WORLD endows DIO with the capability to effectively tackle generative RPM problems.

\subsection{WORLD Guides DIO on Generative RPM problems}
Beyond enabling DIO to solve discriminative RPM tasks, the WORLD method also equips it with the capability to tackle generative RPM problems. This method constructs a complete generation pipeline for DIO: ``estimating the solution distribution $\rightarrow$ sampling the solution distribution $\rightarrow$ decoding the sampling results''.

The WORLD method extends DIO by incorporating an external GMM $\{U_m | m \in [1, M]\}$, an additional loss term $\ell_{\text{Weighted}}$, and a decoder $\mathcal{D}(\cdot)$. Once trained in this framework, DIO can not only discriminate among the options for each context matrix but also generate the corresponding solutions.
Specifically, as shown in Formula~\eqref{bound}, the loss term $\ell_{\text{Weighted}}$ prompts DIO to assign a context-dependent weight to each GMM component $U_m$, where the weight reflects the likelihood that $U_m$ contains the correct options for the given context matrix.
Within this loss term, the weight for each $U_m$ given the context matrix features $\{z_{ij} | i \in [1,8]\}$ is calculated as:
\begin{align}\label{weighted}
    S_{mj} = \tilde{\text{DIO}}(\{z_{ij}, U_m | i \in [1,8]\}).
\end{align}
Since these weights $\{S_{mj}| m \in [1,M], j \in [1,N]\}$ indicate the likelihood that the components cover the correct options, we not only gain the chance to sample diverse incorrect options from low-weight components to tighten the lower bound on the mutual information $I(\{x_i\}_{i = 1}^{8}; x_{\alpha})$, but also have the opportunity to combine high-weight components into a solution distribution to provide promising options.
To this end, we adopt the Gumbel-Max trick~\cite{Gumbel} to sample $N$ high-weight components as the solution distribution that DIO estimates for the context matrix $\{x_i| i\in[1,8]\}$. Specifically, this trick is applied as follows:
\begin{align}
    G_{mj} &= -\log(-\log(\text{U}(0,1))) \label{eq:gumbel} \\
    \tilde{S}_{mj} &= S_{mj} + G_{mj} \label{eq:noisy_logit} \\
    \hat{m}_j &= \underset{m}{\text{argmax}}\,\{\tilde{S}_{mj}|m\in[1,M]\}\label{eq:gmm_sample}
\end{align}
In this case, the sampled indices $\{\hat{m}_j|j\in[1,N]\}$ can point out the solution distribution $\{U_{\hat{m}_j}| j \in [1, N]\}$. Subsequently, Formula~\eqref{generate_ability_d} shows that the well-trained decoder $\mathcal{D}(\cdot)$ is capable of transforming samples from this estimated distribution $\{U_{\hat{m}_j}| j \in [1, N]\}$ into the generated answer $\hat{x}$:
\begin{align}
    &\hat{x}=\mathcal{D}(\{U_{\hat{m}_j}+\epsilon_{ij}|j\in[1,N]\})\label{decode}\\\nonumber
    &\text{where}\quad \epsilon_{ij}\sim \mathcal N(0,1)
\end{align}
In a nutshell, WORLD requires DIO to interact with a GMM that fully encodes RPM image features, a demand that far surpasses traditional discriminative RPM tasks, thereby sparking DIO’s capacity to solve generative RPM problems.


Commendably, WORLD matches the multi-solution characteristic of generative RPM problems. Equations~\eqref{eq:gumbel}–\eqref{eq:gmm_sample} show that the Gumbel-Max trick transfers the diversity inherent in the Gumbel noise \(G_{mj}\) to the solution index \(m_j\), so that the estimated distribution \(\{U_{\hat m_j}| j\in[1,N]\}\) ultimately inherits this diversity. Equation~\eqref{decode} then demonstrates that decoding the samples $\{U_{\hat m_j}+\epsilon_{ij}| j\in[1,N]\}$ has the potential to embed sampling diversity $\epsilon_{ij}$ directly into the resulting solutions $\hat x$. Iterating the estimate $\rightarrow$ sample $\rightarrow$ decode pipeline, WORLD systematically explores multiple valid solutions and fully embraces the multi-solution logic of generative RPM tasks.

\subsection{Hierarchical Interaction and EMA-based Optimization.}

WORLD minimizes the loss function \(\ell_{\text{WORLD}}\) to embed the RPM image features \(u_{ij}+b_{ij}\) encoded by DIO into its trainable GMM \(\{U_m | m\in[1,M]\}\).  
Evidently, the component count \(M\) governs WORLD’s effectiveness: an insufficient \(M\) forces DIO to compress intricate image features into a low-capacity GMM, diluting the expressivity of these features, degrading DIO’s performance, and obscuring WORLD’s true potential. Although increasing \(M\) alleviates these issues, it also incurs computational overhead that scales linearly with \(M\) and quickly drives up the cost of optimizing \(\ell_{\text{WORLD}}\). This places the setting of $M$ in a quandary when WORLD confronts demanding generative RPM problems.

To this end, we propose two dedicated improvements for the two loss terms \(\ell_1\) (i.e., Formula~\eqref{weight_1}) and \(\ell_3\) (i.e., Formula~\eqref{weight_3}) within \(\ell_{\text{WORLD}}\): hierarchical interaction and exponential-moving-average (EMA)-based component optimization.
Specifically, we adopt a G-head hierarchical interaction mechanism that splits the image feature \( u_{ij} \) into \( G \) independent heads \( \{u^{(g)}_{ij} | g \in [1,G]\} \), partitions the \( M \) trainable vectors \( \{U_m | m \in [1,M]\} \) into \( G \) groups \( \{U^{(g)}_m | m \in [1,M], g \in [1,G]\} \), and then performs head-wise hierarchical optimization with the rewritten term \( \ell_1 \) given by:
\begin{align}\label{weight_1_plus}
&\ell_1 = \frac{1}{16N}\sum_{i=1}^{16}\sum_{j=1}^{N}\Bigl\| u_{ij} - \operatorname{concat}\bigl[U^{(g)}_{\hat{m}_g}\bigr]_{g=1}^{G}\Bigr\|_2^2\\
&\quad\quad\text{where}\quad \hat{m}_g = \underset{m\in[1,M]}{\text{argmin}} \bigl\| u^{(g)}_{ij} - U^{(g)}_m \bigr\|_2^2.\nonumber
\end{align}
This enables the same $M$ trainable vectors to represent a GMM with $M^G$ components, whereby the capacity grows exponentially in $G$ instead of linearly in $M$, eliminating the need to drastically increase $M$ for greater expressivity.
However, symmetrically extending the loss term \(\ell_3\) in the same fashion would significantly increase optimization overhead. Therefore, rather than supervising the hierarchical training of components $\{U^{(g)}_m | m \in [1,M], g \in [1,G]\}$ with $\ell_3$, we switch to an EMA-based update strategy that enables these GMM components to evolve without the gradient back-propagation overhead. The EMA-based update for $U_{m}^{(g)}$ is given by:
\begin{align}
\hat{m}_{ij}^{(g)}
&=\underset{m\in[1,M]}{\mathrm{argmin}}
\|u_{ij}^{(g)}-U_{m}^{(g)}\|_2^2\label{count}\\      
n_{m}^{(g)}
&=\sum_{i,j}\mathcal{I}[m=\hat{m}_{ij}^{(g)}],
s_{m}^{(g)}
=\sum_{i,j}\mathcal{I}[m=\hat{m}_{ij}^{(g)}]\frac{u_{ij}^{(g)}}{n_{m}^{(g)}}\label{totu_count}\\
N_{m}^{(g)}
&\leftarrow\eta N_{m}^{(g)}+\gamma \,n_{m}^{(g)},\,
S_{m}^{(g)}
\leftarrow\eta S_{m}^{(g)}+\gamma\,s_{m}^{(g)}\label{moment}\\  
U_{m}^{(g)}
&\leftarrow\dfrac{S_{m}^{(g)}}{N_{m}^{(g)}+\varepsilon}
\Bigg/\sum_{k=1}^{M}\dfrac{N_{k}^{(g)}}{N_{k}^{(g)}+\varepsilon}\label{moment_U}  
\end{align}
Here, $\eta$ is 0.99; $\gamma$ is 0.02; $\varepsilon$ is $10^{-5}$; $N_{m}^{(g)}$ and $S_{m}^{(g)}$ are initialized to 0; \(\mathcal{I}[\cdot]\) denotes the indicator function.

Equations \eqref{count} and \eqref{totu_count} compute the count \(n_{m}^{(g)}\) and the mean \(s_{m}^{(g)}\) of the representations \(u_{ij}^{(g)}\) assigned to the GMM component \(U_{m}^{(g)}\), while Equations \eqref{moment} and \eqref{moment_U} perform the momentum updates of \(n_{m}^{(g)}\), \(s_{m}^{(g)}\), and \(U_{m}^{(g)}\). These updates are executed per gradient step on the \(\ell_1\) loss term, ensuring that the GMM evolves in lockstep with DIO.

\subsection{EMA-Based Efficient Estimation of Solution Distribution}
The G-head hierarchical interaction mechanism enables \(M\) vectors to support a GMM with \(M^{G}\) components; however, as indicated by Equations \eqref{weighted}–\eqref{eq:gmm_sample}, it causes a sharp increase in the number of times \(\tilde{\text{DIO}}\) must weigh the components, imposing a heavy computational burden.

To this end, a EMA-Based approach is adopted: cluster the $M^G$ components into $M$ centroids, assign weights to these centroids using the procedure in Equation \eqref{weighted}, and perform Gumbel-MAX sampling on the $M$ weights as indicated by Equation \eqref{eq:gmm_sample}; then iteratively repeat clustering, weight assignment, and sampling within the selected cluster until only $M$ components remain. Finally, use Gumbel-MAX to select the component $U_{\hat{m}}$ that can serve as the solution distribution from these $M$ components. This reduces the number of weighting operations that $\tilde{\text{DIO}}$ must perform from $M^G$ to $M \times G$.

This recursive scheme demands that $\tilde{\text{DIO}}$ be Lipschitz continuous (i.e., it must assign similar weights to similar components). We therefore apply spectral normalization \cite{Lipschitz} to every parameter matrix in $\tilde{\text{DIO}}$ and apply Lipschitz normalization \cite{Lipschitz_attention} to the self-attention mechanism within each block, ensuring the required Lipschitz property.

\subsection{Validation of Generative RPM problems}
Notably, the validation of generative RPM has long been hindered by its multi-solution nature: any answer that satisfies the progressive pattern can be correct, rendering traditional point-to-point comparisons ineffective. Validation must therefore adopt the same progressive-pattern-driven principle, one that is already embedded in the DIO model trained with the $\ell_{\text{DIEGO}}$ loss, enabling it to perform verification effectively.


\section{Experiment}
All experiments in this paper were conducted on four A100-80G GPUs using Python and the PyTorch\cite{Pytorch} framework.
This paper proposes four configurations for RPM problems:
    \subsubsection{Solo DIO} 
    In the DIO model, the hyperparameter \( N \) is set to 16, indicating that the ViT within its image feature extraction module employs a \( 4 \times 4 \) patch division strategy;

    \subsubsection{Brando-enhanced DIO} 
    In the Brando framework, the hyperparameter $\beta$ is set to 24, which means the Brando network provides $8$ ($\beta - 16 = 8$) constructive hypothetical incorrect options to the DIO in each iteration;

    \subsubsection{WORLD-augmented DIO} 
    In the WORLD framework, when solving discriminative RPM problems, \(M\) is set to $2^{11}$, and a random window of size 256 is defined to select the components used for computing the loss term \(\ell_{\text{Weighted}}\). If the \(G\)-head hierarchical interaction is activated, \(G\) is set to 2, \(M\) remains $2^{11}$, and each head employs a window of size 16 to compute the \(\ell_{\text{Weighted}}\) loss. When the framework is applied to generative RPM tasks, \(G\) is set to 2, and \(M\) is set to $2^{14}$; the random window for \(\ell_{\text{Weighted}}\) inside each head is kept at 16.
    
    \subsubsection{DIEGO-guided DIO} 
    In the {DIEGO} framework, the parameter \( F \) is determined by the number of metadata modalities in the targeted {RPM} problem. 

This paper conducts experiments on the RAVEN \cite{RAVENdataset}, I-RAVEN \cite{I-RAVEN}, and PGM \cite{PGMdataset} datasets, centered around these four configurations. To ensure a fair comparison, we use the same optimizer hyperparameters, dataset scale, and batch size as those of the current SOTA RS-Tran model \cite{RS}.

\subsection{Experiment on RAVEN and I-RAVEN Problems}

Table \ref{RAVEN_IRAVEN_Results} summarizes the performance of our four configurations in comparison with previous models on the RAVEN and I-RAVEN datasets, where ``WORLD$^G$" denotes the activation of the $G$-head hierarchical interaction.
\begin{table}[h]
\caption{Reasoning Accuracies on RAVEN and I-RAVEN.}
\label{RAVEN_IRAVEN_Results}
\centering
\resizebox{\linewidth}{!}{
\begin{tabular}{cccccccccc}
\toprule
\toprule
&\multicolumn{8}{c}{Test Accuracy(\%)}& \\
\cmidrule{2-9}
Model&Average&Center&2$\times$2 Grid&3$\times$3 Grid&L-R&U-D&O-IC&O-IG \\
\midrule
{CoPINet\cite{CoPINet}}&52.96/22.84&49.45/24.50&61.55/31.10&52.15/25.35&68.10/20.60&65.40/19.85&39.55/19.00&34.55/19.45 \\
\cmidrule{2-9}
{PrAE Learner\cite{PrAE}}&65.03/77.02&76.50/90.45&78.60/85.35&28.55/45.60&90.05/96.25&90.85/97.35&48.05/63.45&42.60/60.70 \\
\cmidrule{2-9}
SAVIR-T \cite{SAVIR-T}&94.0/98.1&97.8/99.5&94.7/98.1&83.8/93.8&97.8/99.6&98.2/99.1&97.6/99.5&88.0/97.2\\
\cmidrule{2-9}
MRNet \cite{MRNet}&96.6/-&-/-&-/-&-/-&-/-&-/-&-/-&-/-\\
\cmidrule{2-9}
RS-TRAN\cite{RS}&{98.4}/98.7&99.8/{100.0}&{99.7}/{99.3}&{95.4}/96.7&99.2/{100.0}&{99.4}/99.7&{99.9}/99.9&{95.4}/95.4 \\
\midrule
DIO & 98.8/99.2 & 100.0/100.0 & 99.6/99.7 & 96.3/97.4 & 99.8/99.9 & 99.8/99.9 & 99.8/99.8 & 96.7/97.9 \\
\cmidrule{2-9}
DIO+Brando&99.2/{99.5}&100.0/{100.0}&99.8/{99.8}&97.4/{98.6}&99.9/99.9&99.9/100.0&100.0/100.0&97.8/{98.8} \\
\cmidrule{2-9}
DIO+WORLD & 99.4/99.7 & 100.0/100.0 & 99.8/99.8 & 98.5/99.2 & 99.9/100.0 & 100.0/100.0 & 100.0/100.0 & 98.4/99.3 \\
\cmidrule{2-9}
DIO+WORLD$^G$ & 99.6/99.7 & 100.0/100.0 & 99.8/99.8 & 99.2/99.5 & 100.0/100.0 & 100.0/100.0 & 100.0/100.0 & 99.3/99.5 \\
\cmidrule{2-9}
DIO+DIEGO&99.9/99.9&100.0/{100.0}&99.9/{99.9}&99.7/99.8&100.0/{100.00}&100.0/{100.0}&100.0/100.0&99.8/99.8 \\
\bottomrule
\bottomrule
\end{tabular}
}
\end{table}
These results indicate that all four configurations outperform the existing baselines on both the RAVEN and I-RAVEN datasets. Notably, the {DIO+DIEGO} configuration achieves a {peak accuracy of 99.9\%}, delivering {significant performance gains} over the prior SOTA model (RS-TRAN\cite{RS}). Particularly in complex reasoning tasks such as {3$\times$3 Grid} and {O-IG}, the DIO series models demonstrate {distinct advantages}, validating the {generalizability and robustness} of their architectural design for abstract visual reasoning challenges.

Moreover, our experiments demonstrate that when no additional causal supervisory signals are introduced, the strength of the mutual information inferred by the reasoning model becomes the primary determinant of its problem-solving performance, offering crucial architectural insights for future abstract-reasoning systems. When progressive pattern descriptions are employed as supervisory signals, a supervision strategy that combines causal-chain modeling with the correction of key chain segments proves to be highly effective. This overturns the long-standing conclusion of prior studies \cite{MRNet,PrAE,SAVIR-T,RS} that direct progressive-pattern supervision inevitably undermines model performance.

\subsection{Experiment on  PGM Problem}
Column 1 of Table \ref{PGM_Results} summarizes the performance comparison among the four DIO configurations and existing methods on the i.i.d. sub-task ``neutral'' of the PGM dataset.
\begin{table}[htbp]
\caption{Accuracies on PGM Generalization Tasks.}
\label{PGM_Results}
\centering
\resizebox{\linewidth}{!}{
\begin{tabular}{lcccccccc}
\toprule
\multirow{2}{*}{Model} & \multicolumn{8}{c}{Dataset} \\
\cmidrule{2-9}
 & Neutral & Interpolation & Extrapolation & \makecell{Held-out\\Attribute\\shape-colour} & \makecell{Held-out\\Attribute\\line-type} & \makecell{Held-out\\Triples} & \makecell{Held-out\\Pairs of\\Triples} & \makecell{Held-out\\Attribute\\Pairs} \\
 \midrule
{CoPINet\cite{CoPINet}}&56.4&-&-&-&-&-&-&-\\
\midrule
SAVIR-T \cite{SAVIR-T}&91.2&-&-&-&-&-&-&-\\
\midrule
RS-CNN\cite{RS}&82.8&-&-&-&-&-&-&-\\
\midrule
MRNet\cite{MRNet} & 94.5 & {68.1} & 19.2 & 16.9 & 30.1 & 25.9 & {55.3} & {38.4} \\
\midrule
RS-Tran\cite{RS} & 97.5 & {77.2} & 19.2 & 12.9 & 24.7 & 22.2 & {43.6} & {28.4} \\
\midrule
\midrule
DIO & 98.1 & 82.0 & 18.8 & 13.6 & 26.5 & 23.3 & 45.5 & 29.9 \\
\midrule
DIO+Brando & 99.0 & 83.2 & 18.2 & 14.2 & 26.7 & 23.6 & 46.8& 32.8 \\
\midrule
DIO+WORLD & 99.3 & {85.4} & 18.9 & 15.0 & 27.6 & 23.8 & {55.5} & 39.0 \\
\midrule
DIO+WORLD$^G$ & \underline{99.5} & \underline{87.8} & 18.9 & 15.0 & 27.6 & 23.8 & \underline{56.2} & 39.5 \\
\midrule
DIO+DIEGO & \textbf{99.6} & \textbf{93.2} & 13.8 & 14.4 & 27.2 & 28.8 & \textbf{98.3} & \textbf{98.3} \\
\bottomrule
\end{tabular}
}
\end{table}
As shown in this column, DIO attains SOTA performance on PGM-neutral, a benchmark on which earlier models achieved relatively low accuracy. Notably, incorporating mutual information lower bound tightening methods (Brando and WORLD) yields substantial improvements for DIO. Moreover, DIO+DIEGO reaches a peak accuracy of 99.6\%, challenging the prevailing conclusion that direct metadata supervision necessarily impairs relational reasoning.

Columns 2 onward in Table \ref{PGM_Results} record the experimental results on the o.o.d. sub-tasks of PGM.
All results in Table~\ref{PGM_Results} jointly show that mutual-information tightening and causal-chain modeling yield comparable gains for DIO on i.i.d. RPM tasks, while the latter delivers significantly better generalization under o.o.d. conditions.  
This suggests that, although mutual information as a statistical measure does not inherently encode the causal relationships vital for reasoning, it remains highly effective for i.i.d. tasks; however, when tackling o.o.d. challenges, adopting causal-chain modeling as the core methodology is the more robust choice. 
We suggest that this occurs because, under i.i.d. conditions, the correlations between contexts and solutions coincide with their causal relationships, allowing mutual-information-driven models to remain effective; in contrast, o.o.d. reasoning tasks expose a divergence between correlation and causality, making causal-chain-driven approaches essential.

\subsection{Experiment on Generative RAVEN and PGM Problems}

We argue that a progressive-pattern-driven discriminative model is a reliable way to verify solutions from generative RPM tasks. Therefore, we use DIO+DIEGO to evaluate the generation quality of DIO+WORLD$^G$. Before running this pipeline, we must report DIEGO’s discriminative performance for progressive patterns under RPM metadata supervision to confirm its reliability as a solution validator. Table \ref{Diego} lists DIEGO's reasoning accuracy for progressive patterns on RAVEN, I-RAVEN, and PGM.
\begin{table}[htbp]
\centering
\caption{DIEGO's Reasoning Accuracy for Progressive Patterns on the RAVEN, I-RAVEN, and PGM Datasets.}\label{Diego}
\renewcommand{\arraystretch}{1} 
\begin{tabular}{cccc}
\hline
 \hline
{Dataset} & {Sub-task} & \multicolumn{2}{c}{{Accuracy(\%)} } \\ 
\hline
\multirow{7}{*}{RAVEN} 
 & center & \multicolumn{2}{c}{entire:99.9} \\ 
 & 2$\times$2 Grid & \multicolumn{2}{c}{entire:99.9} \\ 
 & 3$\times$3 Grid & \multicolumn{2}{c}{entire:99.8} \\ 
 & L-R & left:99.9& right:99.9 \\ 
 & U-D & up:99.9& down:99.9  \\ 
 & OI-C & in:99.9& out:99.9 \\ 
 & OIG & in:99.8& out:99.9  \\
 \hline
\multirow{7}{*}{I-RAVEN} 
 & center & \multicolumn{2}{c}{entire:99.9} \\ 
 & 2$\times$2 Grid & \multicolumn{2}{c}{entire:99.9} \\ 
 & 3$\times$3 Grid & \multicolumn{2}{c}{entire:99.8} \\ 
 & L-R & left:99.9& right:99.9  \\ 
 & U-D & up:99.9& down:99.9 \\ 
 & OI-C & in:99.9& out:99.9\\ 
 & OIG & in:99.8& out:99.9 \\
 \hline
PGM & neutral & shape:99.7 & line:99.9\\ 
 \hline
  \hline
\end{tabular}
\end{table}

\begin{figure}[h]\centering
	\includegraphics[trim=0cm 0cm 0cm 0cm, clip, width=8.5
 cm]{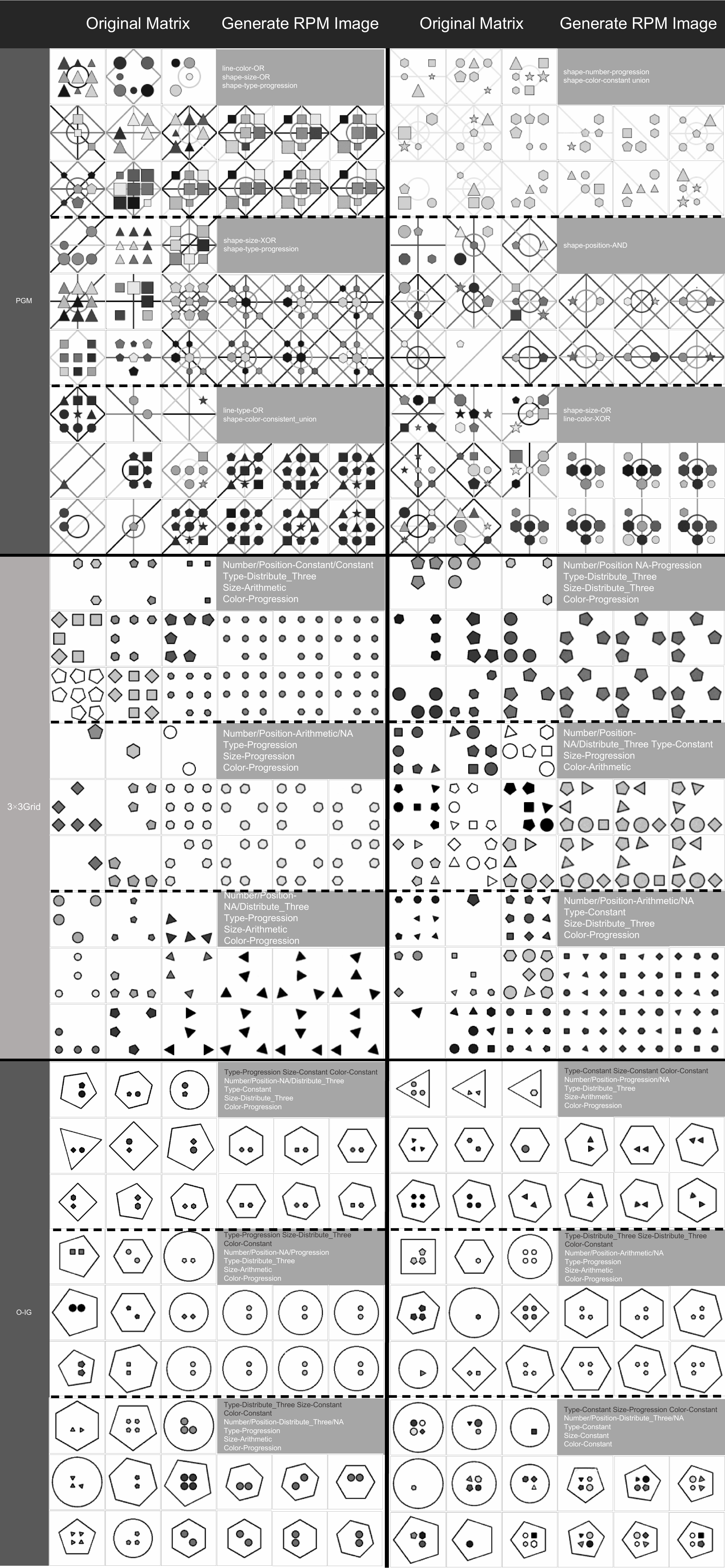}
	\caption{Generated RPM images.}
\label{DIO_Generate_ALL}
\end{figure}

Table \ref{Diego} shows that DIO+DIEGO achieves extremely high accuracy on progressive-pattern reasoning, making it a suitable validator for generative RPM tasks. Consequently, we use DIO+DIEGO to evaluate DIO+WORLD$^G$ and report its accuracy on generative RAVEN, I-RAVEN, and PGM-neutral problems in Table \ref{Combined_Results_Simplified_}.
\begin{table}[htbp]
\caption{Generation Accuracies of DIO+WORLD$^G$}
\label{Combined_Results_Simplified_}
\centering
\begin{tabular}{lccccc}
\toprule
{Dataset} & {Sub-task} & {Accuracy (\%)} & Average (\%)\\
\midrule
\multirow{7}{*}{\makecell{RAVEN/\\I-RAVEN}} & Center & 98.0/98.2 &\multirow{7}{*}{\makecell{91.3/92.6 }}\\
 & 2$\times$2 Grid & 82.2/82.0 \\
 & 3$\times$3 Grid & 81.5/81.8 \\
 & L-R & 96.5/96.9 \\
 & U-D & 95.7/96.9 \\
 & O-IC & 94.9/96.5 \\
 & O-IG & 90.5/96.2 \\
\midrule
PGM & Neutral & {80.5} &80.5\\
\bottomrule
\end{tabular}
\end{table}
The relevant generation results are provided in Figure \ref{DIO_Generate_ALL}, where, for RAVEN, we display only the most typical 3$\times$3 Grid and O-IG sub-task outcomes.

Table \ref{Combined_Results_Simplified_} shows that, without extra supervisory signals, DIO+WORLD$^G$ has attained competitive results in open-ended generative RPM problems, strongly confirming its significant research value.
%


\subsection{Semantic Disentanglement via WORLD Encoding.}

\begin{figure}[h]\centering
	\includegraphics[trim=0cm 6.6cm 0cm 0cm, clip, width=8.5
 cm]{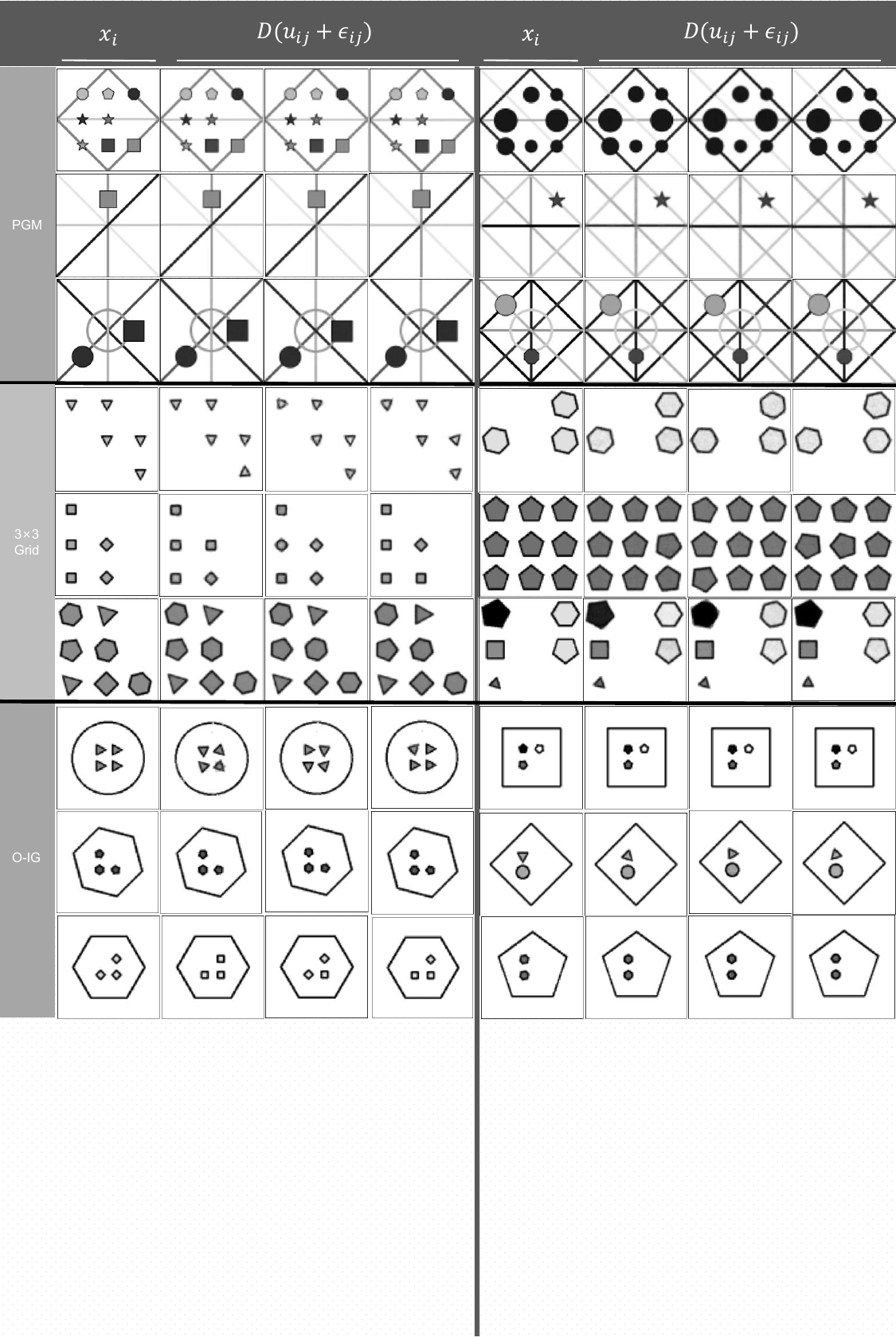}
	\caption{Verification on Image Semantics.}
\label{DIO_generate_layout}
\end{figure}

Under the WORLD framework, as illustrated in Figure \ref{The modified Image Feature Extraction Module.}, each RPM image \(x_i\) is encoded into two latent features: \(u_{ij}\) and \(b_{ij}\sim \mathcal N(0,1)\), and the downstream reasoning operates on \(u_{ij}+\epsilon_{ij}\). This may give rise to a latent division of labor: \(u_{ij}\) encapsulates the core semantics required for reasoning, while the decoder \(\mathcal D(\cdot)\), trained to maximize the mutual information between \(u_{ij}+b_{ij}\) and \(x_i\), pushes \(b_{ij}\) to carry task-irrelevant perturbations. To verify this hypothesis, we compare the reconstructed image $\mathcal D(u_{ij}+\epsilon_{ij})$ with the original $x_i$ in terms of semantics, revealing how random variations in $b_{ij}$ drive semantic drift. The comparisons are shown in Figure~\ref{DIO_generate_layout}.

As can be seen in Figure~\ref{DIO_generate_layout}, DIO+WORLD$^G$ begins to disentangle perturbing semantics, such as the occlusion relationships among background lines in PGM and the rotation angles of entities in RAVEN, from $u_{ij}$ to some extent.

\section{Conclusion}

This study has constructed a comprehensive abstract reasoning solution through systematic innovations, with four core innovations forming a progressive technical system that achieves significant advancements in both theoretical breakthroughs and empirical performance:

\subsubsection{DIO} We propose that structuring networks around causal chains can enhance reasoning performance. We further suggest that the learning objectives of end-to-end reasoning models, including DIO, are essentially variational lower bounds on the mutual information between contexts and correct options. Consequently, however elaborate the network, performance hinges on the tightness of this bound. This points out that the key to machine intelligence lies in mutual-information tightening or causal-chain modeling.
 
\subsubsection{Brando} We have integrated the concept of mutual information tightening into abstract reasoning frameworks, showing that training with more constructive incorrect options serves as a key mechanism for bounding mutual information. This explores novel pathways for variational inference methodologies. Concurrently, we have developed a constructive hypothetical incorrect option mapping mechanism to analytically  constrain the variational lower bound of mutual information.

\subsubsection{WORLD} We construct a GMM for the distribution of RPM image features, enabling targeted sampling of infinite and semantically rich options. This approach not only helps the DIO model tighten mutual information and, to some extent, disentangle perturbing semantics, but also equips it with the capability to solve open-ended generative RPM problems.
 
\subsubsection{DIEGO} We introduce a non-statistical constraint that precisely corrects semantic biases in the causal chain modeled by DIO, and we establish mappings between the model's reasoning chains and the ground-truth causal chains, aligning internal representations with human cognitive frameworks.

This work unites four causal-information innovations that together yield a complete abstract-reasoning stack.



\newpage


\begin{thebibliography}{1}
\bibliographystyle{IEEEtran}


\bibitem{GAN} Goodfellow, I. \emph{et~al.} Generative adversarial networks. Communications of the ACM, 63(11), 139-144 (2020).

\bibitem{VAE} Kingma, D. P., \& Welling, M. Auto-encoding variational bayes. Preprint at https://arxiv.org/abs/1312.6114 (2014).


\bibitem{Transformer} Vaswani, A. \emph{et~al.} Attention is All You Need. In Advances in Neural Information Processing Systems, (2017).

\bibitem{ResNet} He, K., Zhang, X., Ren, S., \& Sun, J. Deep Residual Learning for Image Recognition. In IEEE Conference on Computer Vision and Pattern Recognition, 770-778 (2016).

\bibitem{A survey of convolutional neural networks}Li, Z., Liu, F., Yang, W., Peng, S., \& Zhou, J. (2021). A survey of convolutional neural networks: analysis, applications, and prospects. IEEE transactions on neural networks and learning systems, 33(12), 6999-7019.

\bibitem{A survey of visual transformers}Liu, Yang, et al. "A survey of visual transformers." IEEE Transactions on Neural Networks and Learning Systems (2023).



\bibitem{GPT-3} Brown, T. \emph{et~al.} Language Models are Few-shot Learners. In Advances in Neural Information Processing Systems, 1877-1901 (2020).


\bibitem{Attention in natural language processing}Galassi, A., Lippi, M., \& Torroni, P. (2020). Attention in natural language processing. IEEE transactions on neural networks and learning systems, 32(10), 4291-4308

\bibitem{survey of natural language processing}Otter, D. W., Medina, J. R., \& Kalita, J. K. (2020). A survey of the usages of deep learning for natural language processing. IEEE transactions on neural networks and learning systems, 32(2), 604-624.


\bibitem{VQA} Antol, S., Agrawal, A., Lu, J., Mitchell, M., Batra, D., Zitnick, C. L., \& Parikh, D. VQA: Visual question answering. In IEEE International Conference on Computer Vision, 2425-2433 (2015).




\bibitem{LLM2}B. Li, Y. Zhang, D. Guo, R. Zhang, F. Li, H. Zhang, K. Zhang, P. Zhang, Y. Li, Z. Liu, and C. Li, "LLaVA-OneVision: Easy Visual Task Transfer," arXiv Preprint arXiv:2408.03326, 2024.


\bibitem{LLM3}T. Zheng, J. Cheng, C. Li, H. Shi, Z. Wang, J. Bai, Y. Song, G. Y. Wong, and S. See, "LogiDynamics: Unraveling the Dynamics of Logical Inference in Large Language Model Reasoning," arXiv Preprint arXiv:2502.11176, 2025.


\bibitem{LLM4}Zhang, Yizhe, et al. "How Far Are We from Intelligent Visual Deductive Reasoning?." arxiv preprint arxiv:2403.04732 (2024).



\bibitem{RPM} Raven J. C. Raven's Progressive Matrices. (Western Psychological Services, (1938).

\bibitem{RAVENdataset} Zhang, C., Gao, F., Jia, B., Zhu, Y., \& Zhu, S. C. Raven: A Dataset for Relational and Analogical Visual Reasoning. In Proceedings of the IEEE/CVF Conference on Computer Vision and Pattern Recognition, 5317--5327 (2019).

\bibitem{PGMdataset} Barrett, D., Hill, F., Santoro, A., Morcos, A., \& Lillicrap, T. Measuring Abstract Reasoning in Neural Networks. In International Conference on Machine Learning, 511-520 (2018).



\bibitem{ViT} Dosovitskiy, A. \emph{et~al.} An Image is Worth 16x16 Words: Transformers for Image Recognition at Scale. Preprint at https://arxiv.org/abs/2010.11929 (2020).


\bibitem{CoPINet} Zhang, C., Jia, B., Gao, F., Zhu, Y., Lu, H., \& Zhu, S. C. Learning Perceptual Inference by Contrasting. In Proceedings of Advances in Neural Information Processing Systems, (2019).

\bibitem{DCNet} Zhuo, T., \& Kankanhalli, M. Effective Abstract Reasoning with Dual-Contrast Network. In Proceedings of International Conference on Learning Representations, (2020).

\bibitem{MRNet} Benny, Y., Pekar, N., \& Wolf, L. Scale-Localized Abstract Reasoning. In Proceedings of the IEEE/CVF Conference on Computer Vision and Pattern Recognition, 12557-12565, (2021).


\bibitem{NCD} Zhuo, Tao and Huang, Qiang \& Kankanhalli, Mohan. Unsupervised abstract reasoning for raven's problem matrices. IEEE Transactions on Image Processing, 8332--8341, (2021).

\bibitem{SAVIR-T} Sahu, P., Basioti, K., \& Pavlovic, V. SAViR-T: Spatially Attentive Visual Reasoning with Transformers. Preprint at https://arxiv.org/abs/2206.09265 (2022).


\bibitem{PrAE} Zhang, C., Jia, B., Zhu, S. C., \& Zhu, Y. Abstract Spatial-Temporal Reasoning via Probabilistic Abduction and Execution. In Proceedings of the IEEE/CVF Conference on Computer Vision and Pattern Recognition, 9736-9746 (2021).

\bibitem{ALANS} Zhang, C., Xie, S., Jia, B., Wu, Y. N., Zhu, S. C., \& Zhu, Y. Learning Algebraic Representation for Systematic Generalization. In Proceedings of the European Conference on Computer Vision, (2022).

\bibitem{NVSA} Hersche, M., Zeqiri, M., Benini, L., Sebastian, A., \& Rahimi, A. A Neuro-vector-symbolic Architecture for Solving Raven's Progressive Matrices. Preprint at https://arxiv.org/abs/2203.04571 (2022).


\bibitem{RS} Wei, Q., Chen, D., \& Yuan, B.  Multi-viewpoint and multi-evaluation with felicitous inductive bias boost machine abstract reasoning ability, IEEE Transactions on Image Processing, vol. 34, pp. 667–677, Jan. 2025.

\bibitem{CRAB}Shi, Fan, Bin Li, and Xangyang Xue. "Abstracting Concept-Changing Rules for Solving Raven's Progressive Matrix Problems." arxiv preprint arxiv:2307.07734 (2023).


  










\bibitem{I-RAVEN} Hu, S., Ma, Y., Liu, X., Wei, Y., \& Bai, S. Stratified Rule-Aware Network for Abstract Visual Reasoning. In Proceedings of the AAAI Conference on Artificial Intelligence, 1567-1574 (2021).


\bibitem{RPMInductivebias} Carpenter, P. A., Just, M. A., \& Shell, P. What One Intelligence Test Measures: a Theoretical Account of the Processing in the Raven Progressive Matrices Test. Psychological review, 97(3), 404, (1990).


\bibitem{InfoNCE} Oord, A. V. D., Li, Y., \& Vinyals, O. Representation Learning with Contrastive Predictive Coding. Preprint at https://arxiv.org/abs/1807.03748 (2019).


\bibitem{infin_atten} Munkhdalai, Tsendsuren, Manaal Faruqui, and Siddharth Gopal. "Leave no context behind: Efficient infinite context transformers with infini-attention." arxiv preprint arxiv:2404.07143 101 (2024).


\bibitem{GMM} {D. A. Reynolds, "Gaussian mixture models," in Encyclopedia of Biometrics, vol. 741, pp. 659-663, 2009, Art. no.}




\bibitem{Causal inference} Pearl, Judea. "Causal inference in statistics: An overview." (2009): 96-146.




\bibitem{Gumbel}Jang, Eric, Shixiang Gu, and Ben Poole. "Categorical reparameterization with gumbel-softmax." arxiv preprint arxiv:1611.01144 (2016).


\bibitem{Lipschitz} Miyato, T., Kataoka, T., Koyama, M., and Yoshida, Y. Spectral normalization for generative adversarial networks. In International Conference on Learning Representations, 2018.

\bibitem{Lipschitz_attention} Dasoulas, George, Kevin Scaman, and Aladin Virmaux. "Lipschitz normalization for self-attention layers with application to graph neural networks." International conference on machine learning. PMLR, 2021.



\bibitem{Pytorch} Paszke, A. \emph{et~al.} Automatic Differentiation in Pytorch. In NIPS Autodiff Workshop, (2017).




\end{thebibliography}
\end{document}


\title{Supplementary Materials for DIO: Refining Mutual Information and Causal Chain to Enhance Machine Abstract Reasoning Ability}

\maketitle

\section{Ablation Study}
This section presents a series of ablation experiments.

\subsection{Ablation Study on DIO}

The ablation study on DIO centers on the design of the network’s feed-forward structure. Within its progressive pattern induction module, DIO employs a permutation and combination strategy ($C_n^r$) to extract progressive patterns from row and column features, aiming to model the causal-chain details inherent in RPM problems.
For ablation, we remove the ($C_n^r$) operation and instead feed all six row-column features directly into the progressive-pattern extractor. By masking the structure that aligns with the causal chain, we assess the importance of causal-chain modeling in network design. The resulting accuracies on key sub-tasks of RAVEN, I-RAVEN and PGM are reported in Table~\ref{Combined_Results_Simplified}.
\begin{table}[htbp]
\caption{Reasoning Accuracies of masked DIO.}
\label{Combined_Results_Simplified}
\centering
\begin{tabular}{lccccc}
\toprule
\multirow{2}{*}{Dataset} &  \multirow{2}{*}{Sub-task}  &\multicolumn{2}{c}{Accuracy (\%)}\\
\cmidrule{3-4}
 &  & Masked DIO& DIO \\
\midrule
\multirow{2}{*}{\makecell{RAVEN/I-RAVEN}} 
 & 3$\times$3 Grid & 95.8/96.9 &96.3/97.4\\
 & O-IG & 96.0/96.1 &96.7/97.9\\
\midrule
PGM & Neutral & {97.2} &98.1\\
\bottomrule
\end{tabular}
\end{table}
Table~\ref{Combined_Results_Simplified} shows that building RPM-solving models that explicitly incorporate causal chains is crucial for superior reasoning performance.

Moreover, across all RAVEN and PGM sub-tasks, DIO uniformly splits every image into 4$\times$4 patches. This design deliberately bypasses the entity layouts in RAVEN’s 2$\times$2 Grid and 3$\times$3 Grid sub-tasks, preventing unintended inductive biases and prior injection while embracing the causal chain rationale of RPM problems: multiple entities must be analyzed holistically. We therefore conducted ablation experiments on the hyperparameter $N$ to examine how aligning the patch grid with RAVEN's 2$\times$2 Grid and 3$\times$3 Grid sub-tasks affects performance. The results are shown in Table \ref{segmentation}. 
\begin{table}[h]
\caption{The accuracy of DIO when its patch grid is aligned with entity positions in the ``2$\times$2 Grid" and ``3$\times$3 Grid".}
\label{segmentation}
\centering
\begin{tabular}{cccc}
\toprule
&\multicolumn{2}{c}{ Accuracy(\%)}& \\
\cmidrule{2-3}
Patch grid& 2$\times$2 Grid&3$\times$3 Grid \\
\midrule
Aligned with entity&95.0/98.3&92.8/94.2\\
\midrule
 4$\times$4 Grid&99.6/99.7 &96.1/97.4\\
\bottomrule
\end{tabular}
\end{table}
These results indicate that favoring priors over the causal chain may undermine model performance.

Finally, we conduct ablation experiments on the loss function of DIO; its expression in the main text is:
\begin{align}\label{loss}
{\ell}_\text{DIO} = \mathbb{E}\left[\log (\frac{\sum_{\substack{{\hat \alpha} = 9\\{\hat \alpha}\neq \alpha}}^{16} \hat{P}( x_{\hat{\alpha}} | \{x_i\}_{i=1}^8, \theta)}{\hat{P}(x_{\alpha} | \{x_i\}_{i=1}^8, \theta)} + \delta)
\right]-\log \delta
\end{align}
This expression is only the definition; it can be simplified as follows:
\begin{align}
\ell_{\text{DIO}}
&= \mathbb{E}\!\left[\log\!\Bigl(\frac{\sum_{\substack{{\hat \alpha} = 9\\{\hat \alpha}\neq \alpha}}^{16}\hat P_{\hat\alpha}}{\hat P_{\alpha}}+\delta\Bigr)\right]-\log \delta \nonumber\\
&= \mathbb{E}\!\left[\log\!\Bigl(\frac{1-\hat P_{\alpha}}{\hat P_{\alpha}}+\delta\Bigr)\right]-\log\delta\nonumber\\
&= \mathbb{E}\!\left[\log\!\Bigl(\frac{1-\hat P_{\alpha}+\delta\hat P_{\alpha}}{\delta\hat P_{\alpha}}\Bigr)\right] \nonumber\\[4pt]
&= \mathbb{E}\!\left[-\log \hat{P}_{\alpha} + \log\!\Bigl(\frac{1-\hat{P}_{\alpha}+\delta\hat{P}_{\alpha}}{\delta}\Bigr)\right]
\end{align}
After simplification, it is evident that \(\ell_{\text{DIO}}\) can be viewed as the traditional cross-entropy (CE) loss augmented with a correction term that accompanies the annealing coefficient $\delta$. Thus, an ablation study on this correction term can begin by letting \(\ell_{\text{DIO}}\) reduce to the CE loss.
\begin{table}[htbp]
\caption{Reasoning Accuracies of DIO+CE.}
\label{Combined_Results_Simplified_}
\centering
\resizebox{\linewidth}{!}{
\begin{tabular}{lccccccc}
\toprule
\multirow{2}{*}{Dataset} &  \multirow{2}{*}{Sub-task}  &\multicolumn{4}{c}{Accuracy (\%)}\\
\cmidrule{3-6}
 &  & DIO+CE & DIO & RS-Tran\cite{RS} & RS-Tran+\(\ell_{\text{DIO}}\)\\
\midrule
\multirow{2}{*}{\makecell{RAVEN/\\I-RAVEN}} 
 & 3$\times$3 Grid & 96.1/97.1 &96.3/97.4 & 95.4/96.7 &95.7/96.9\\
 & O-IG            & 96.4/97.5 &96.7/97.9 & 95.4/95.4 &95.9/96.0\\
\midrule
PGM & Neutral       & {97.8} &98.1 & {97.5} &97.7\\
\bottomrule
\end{tabular}
}
\end{table}
Table \ref{Combined_Results_Simplified_} shows that, on RPM problems, \(\ell_{\text{DIO}}\) slightly outperforms the standard CE loss and can provide the already CE-trained SOTA model RS-Tran with a modest extra boost.

\subsection{Ablation Study on Brando}
The ablation study on Brando focuses on the hyperparameter~$\beta$, which denotes the number ($\beta{-}17$) of hypothetical options that the Brando network provides to DIO.
We conduct ablation experiments on the DIO$+$Brando under different~$\beta$ values across key sub-tasks of RAVEN, I-RAVEN, and PGM, and report the results in Table~\ref{beta_Combined_Results_Simplified}.
\begin{table}[htbp]
\caption{Accuracies of DIO+Brando under Different $\beta$.}
\label{beta_Combined_Results_Simplified}
\centering
\renewcommand{\arraystretch}{1.2} 
\resizebox{\linewidth}{!}{
\begin{tabular}{ccccccc} 
\toprule
\multirow{2}{*}{Dataset}&\multirow{2}{*}{Sub-task}&\multicolumn{5}{c}{Accuracy(\%)}\\\cmidrule(lr){3-7}
& & $\beta$=24(8) & $\beta$=22(6) & $\beta$=20(4) & $\beta$=18(2) & {Solo DIO} \\
\midrule
\multirow{4}{*}{\makecell{RAVEN/\\I-RAVEN}} 
& 3$\times$3 Grid     & 97.4 / 98.6 & 97.0 / 98.1 & 96.9 / 98.1 & 96.6 / 97.9 & 96.3 / 97.4 \\
 & O-IG               & 97.8 / 98.8 & 97.4 / 98.4 & 97.3 / 98.4 & 97.3 / 98.2 & 96.7 / 97.9 \\
 \cmidrule(lr){3-7}
 & 3$\times$3 Grid$^*$ & 97.0 / 98.0 & 96.7 / 97.9 & 96.6 / 97.8 & 96.6 / 97.7 & 96.3 / 97.4 \\
 & O-IG$^*$           & 97.3 / 98.3 & 97.1 / 98.2 & 97.0 / 98.2 & 97.0 / 98.2 & 96.7 / 97.9 \\
\midrule
\multirow{2}{*}{PGM} & Neutral & 99.0 & 98.8 & 98.8 & 98.5 & 98.1 \\
 \cmidrule(lr){3-7}
& Neutral$^*$ & 98.7 & 98.4 & 98.4 & 98.3 & 98.1 \\
\bottomrule
\end{tabular}
}
\end{table}
Table \ref{beta_Combined_Results_Simplified} indicates that even when only two constructive hypothetical options are introduced into DIO, significant performance improvements can be achieved.
The asterisk ($^*$) on a sub-task indicates that when the Brando network is applied to this task, the process of learning historical image features defined by Equation (26) in the main text is deactivated, facilitating the ablation and analysis of this mechanism. The asterisk-marked results demonstrate that learning historical image features is effective in mapping out constructive hypothetical options.

\subsection{Ablation Study on WORLD}
The ablation study on WORLD focuses on the hyperparameter $M$. 
$M$ indicates the number of components in the GMM constructed by {WORLD} for RPM image features. 
The reasoning accuracies of DIO+WORLD on the key sub-tasks of RAVEN, I-RAVEN, and PGM under different $M$ values are recorded in Table \ref{M_Combined_Results_Simplified}.
\begin{table}[htbp]
\caption{Accuracies of DIO+WORLD under Different $M$.}
\label{M_Combined_Results_Simplified}
\centering
\renewcommand{\arraystretch}{1.2}
\resizebox{\linewidth}{!}{
\begin{tabular}{ccccccc}
\toprule
\multirow{4}{*}{Dataset}&
\multirow{4}{*}{Sub-task}&
\multicolumn{5}{c}{Accuracy (\%)}\\
\cmidrule{3-7}
&&\multirowcell{2}{$M$=$2^{14}$,\\$G$=2}&
  \multirowcell{2}{$M$=$2^{11}$,\\$G$=2}&
  \multirowcell{2}{$M$=$2^{10}$,\\$G$=2}&
  \multirowcell{2}{$M$=$2^{6}$,\\$G$=2}&
  \multirow{2}{*}{Solo DIO}\\
&&&&&&\\
\midrule
\multirow{2}{*}{\makecell{RAVEN/\\I-RAVEN}}
& 3$\times$3 Grid & 99.5 / 99.7 & 99.2 / 99.5 & 98.9 / 99.4 & 98.2 / 98.8 & 96.3 / 97.4\\
& O-IG           & 99.6 / 99.7 & 99.3 / 99.5 & 99.0 / 99.3 & 98.5 / 98.9 & 96.7 / 97.9\\
\midrule
PGM & Neutral & 99.5 & 99.5 & 99.5 & 99.0 & 98.1\\
\midrule
\midrule
& & $M$=$2^{14}$ & $M$=$2^{11}$ & $M$=$2^{10}$ & $M$=$2^{6}$ & Solo DIO\\
\midrule
\multirow{2}{*}{\makecell{RAVEN/\\I-RAVEN}}
& 3$\times$3 Grid & 98.5 / 99.2 & 98.5 / 99.2 & 98.0 / 98.8 & 96.5 / 97.9 & 96.3 / 97.4\\
& O-IG           & 98.5 / 99.4 & 98.4 / 99.3 & 98.2 / 98.8 & 97.1 / 98.0 & 96.7 / 97.9\\
\midrule
PGM & Neutral & 99.3 & 99.3 & 99.0 & 97.8 & 98.1\\
\bottomrule
\end{tabular}
}
\end{table}

As shown in Table \ref{M_Combined_Results_Simplified}, when WORLD constructs a few-component GMM ($M$=$64$), it negatively impacts DIO's performance in the PGM task.
This paper contends that this phenomenon arises because a 64-component GMM lacks the capacity to capture the inherent complexity of PGM image features.
If DIO attempts to fit these more intricate features into such a simplified GMM, a performance drop is inevitable.

Experimental results also demonstrate that the proposed \(G\)-head hierarchical interaction mechanism, designed to meet WORLD's demand for a larger number of GMM components, is effective; without increasing \(M\), it achieves superior performance solely by splitting into multiple heads. Even when $M$ is set to more than 2048, using multiple heads ($G$=2) still improves discriminative performance.

Further observation shows that, without the $G$-head hierarchical interaction, increasing $M$ beyond 2048 components yields little gain; however, activating the mechanism can break through this plateau. To fully verify the discriminative capability of DIO+WORLD$^G$, we set $G$ to 16 and $M$ to $2^{11}$ and conduct experiments on the PGM-neutral task; the results are reported in Table \ref{PGM_Results}.
\begin{table}[htbp]
\caption{Reasoning Accuracies on PGM-neutral.}
\label{PGM_Results}
\centering
\begin{tabular}{ccc}
\toprule
Model&Test Accuracy(\%) \\
\midrule
\multirowcell{2}{DIO+WORLD$^G$\\($M$=$2^{11}$, $G$=16)}&\multirowcell{2}{99.6}\\
&\\
\bottomrule
\end{tabular}
\end{table}
Previous RPM solvers seldom broke the 99\% accuracy barrier on the PGM-neutral benchmark. Simply by activating the $G$-head hierarchical interaction mechanism, our DIO+WORLD not only crashes through that ceiling but also keeps posting incremental gains, further confirming the value of this mechanism.